\documentclass[journal]{IEEEtran}
\usepackage{cite}

\ifCLASSINFOpdf
   \usepackage[pdftex]{graphicx}
  \graphicspath{{./images/}}
\else
   \usepackage{graphicx}
   \graphicspath{{./images/}}
   \DeclareGraphicsExtensions{.eps,.pdf,.png,.jpg}
\fi
\usepackage{amsfonts}
\usepackage{amssymb}
\usepackage[cmex10]{amsmath}
\usepackage{multirow}

\makeatletter
\let\MYcaption\@makecaption
\makeatother
\usepackage[font=footnotesize]{subcaption}
\makeatletter
\let\@makecaption\MYcaption
\makeatother

\ifCLASSOPTIONcompsoc
  \usepackage[caption=false,font=normalsize,labelfont=sf,textfont=sf]{subfig}
\else
  \usepackage[caption=false,font=footnotesize]{subfig}
\fi
\newcommand{\norm}[1]{\left|\left|#1\right|\right|}

\newtheorem{theorem}{Theorem}
\newtheorem{lemma}{Lemma}

\makeatletter
\@addtoreset{theorem}{section}
\makeatother

\begin{document}
%
\title{Local Similarities, Global Coding: An Algorithm for Feature Coding and its Applications}
%
%
%

\author{Amirreza~Shaban,~\IEEEmembership{Student Member,~IEEE,}
        Hamid~R.~Rabiee ,~\IEEEmembership{Senior~Member,~IEEE,}
        and~Mahyar~Najibi
\thanks{J. Doe and J. Doe are with Anonymous University.}
\thanks{Manuscript received April 19, 2005; revised December 27, 2012.}}

%
%

\markboth{Journal of \LaTeX\ Class Files,~Vol.~11, No.~4, December~2012}%
{Shell \MakeLowercase{\textit{et al.}}: Bare Demo of IEEEtran.cls for Journals}
%



\maketitle

\begin{abstract}
Data coding as a building block of several image processing algorithms has been received great attention recently. Indeed, the importance of the locality assumption in coding approaches is studied in numerous works and several methods are proposed based on this concept. We probe this assumption and claim that taking the similarity between a data point and a more global set of anchor points does not necessarily weaken the coding method as long as the underlying structure of the anchor points are taken into account. Based on this fact, we propose to capture this underlying structure by assuming a random walker over the anchor points. We show that our method is a fast approximate learning algorithm based on the diffusion map kernel. The experiments on various datasets show that making different state-of-the-art coding algorithms aware of this structure boosts them in different learning tasks.
\end{abstract}

\begin{IEEEkeywords}
Sparse coding, local coordinate coding, diffusion kernel, image classification, image clustering. 
\end{IEEEkeywords}

%
\IEEEpeerreviewmaketitle

\section{Introduction}
%
%
%
%
\subsection{Coding methods}
\IEEEPARstart{F}{eature} coding is of primary importance as a common building block in many different image processing algorithms that extract a high level representation from images in order to transform a nonlinear high dimensional learning problem in the low-level feature space into a simpler one in the coding space \cite{lcc09}.  Coding is used in a wide variety of applications in computer vision, like background modeling \cite{cevher2008compressive,dikmen2008robust}, super resolution \cite{yang2008image}, tracking \cite{liu2011robust,wang2013online} (See \cite{zhang2012sparse} for detailed review), object recognition \cite{ganesh2009robust}, image classification \cite{bow04,llc10,sac10,scspm09} pose estimation \cite{bowpos06} and image annotation \cite{wang2009multi}.

Assuming $\mathbf{x}$ denotes the low level representation of an image and $\mathbf{D}$ is a dictionary matrix composed of $K$ learned visual bases as its columns. The goal of a coding algorithm is to map $\mathbf{x}$ to a $K$ dimensional vector $\mathbf{c}(\mathbf{x})$ where its $i$'th element, $\mathbf{c}_{(i)}(\mathbf{x})$ indicates the affinity of the data point $\mathbf{x}$ to the $i$'th basis in $\mathbf{D}$. Many coding algorithms have recently been proposed with different approaches to compute these affinities, some based on a similarity measurement \cite{sac10} and others on the reconstruction criteria \cite{llc10,scspm09}. In one of the most primitive methods called Vector Quantization (VQ), each feature vector is assigned to the nearest basis which is learned by $k$-means clustering. In this approach, the information loss in representing an image feature is high due to the hard-assignment nature. In Soft Assignment Coding (SAC) \cite{sac10} the visual word ambiguity is modeled by assigning a feature to several bases by the following similarity measure: 
\begin{equation}
\mathbf{c}_{(i)}(\mathbf{x}) = \frac{1}{Z} \exp \Big(- \frac{|| \mathbf{x} - \mathbf{b_i} || ^ 2}{2 \sigma^ 2} \Big)
\end{equation}
where $Z$ is a normalization factor and the bandwidth parameter $\sigma$ controls the softness of the assignment function and leads to the hard-assignment in the extreme case when $\sigma \rightarrow 0$. 

Sparse coding refers to a class of algorithms that find the sparse representations of the low level features. As an unsupervised learning algorithm, given unlabeled input data, it learns an over complete set of bases and coding vectors that capture high-level features in the images. Sparse coding uses a linear combination of small number of bases that best represent the input image through a reconstruction based optimization problem. Authors in \cite{olshausen1997sparse} state that sparse coding method is similar to what is done in the visual cortex when bases resemble the receptive fields. The sparse coding problem can be formulated as follows:
\begin{equation} \label{eq_sparse_opt}
\begin{split}
& \min_{\mathbf{C}} \sum_{\mathbf{x} \in \mathbf{X}} || \mathbf{x} - \mathbf{B} \text{ } \mathbf{c(x)} ||_2 ^ 2 + \lambda || \mathbf{c(x)} ||_p  \\
\end{split}
\end{equation}
where $\lambda$  controls the sparsity of coding schema. To enforce the sparsity constraint, the problem must be solved for $p=0$ which is called the $\ell^0$ optimization. However, In  \cite{davis1997adaptive} it is proven that finding the optimal solution for the $\ell^0$ optimization is NP-hard. To solve the problem approximately many optimization approaches are proposed, some to solve the $\ell^0$ problem by greedy approaches \cite{chen1989orthogonal,mallat1993matching,pati1993orthogonal} and others to convexify the problem by replacing the $\ell^0$ norm with $\ell^1$ \cite{chen1998atomic}. 

For making the dictionary more specific to the dataset, several methods have been proposed to learn it \cite{aharon2006img, lewicki2000learning} instead of using predefined dictionaries. Recently, supervised dictionary learning methods \cite{mairal2012task, zhang2010discriminative} have also been proposed to increase the discriminant power of the sparse coding algorithm. Sparse coding and its variations \cite{zheng2011graph, gao2013laplacian}  have successfully applied to many computer vision tasks like image denoising \cite{elad2006image}, image classification \cite{scspm09} and face recognition \cite{ganesh2009robust}.

Regularized sparse coding \cite{zheng2011graph} is an extension of sparse coding proposed to cope with data having a manifold structure that enforces smooth variation of the sparse codes with respect to the manifold structure of the data points by adding a graph regularization term $\sum_{\mathbf{x,y} \in \mathbf{X}} (\mathbf{c(x)} - \mathbf{c(y)})^2 \mathbf{W_{xy}}$ to the sparse coding optimization function where $\mathbf{W_{xy}}$ reflects the local similarity of data points $\mathbf{x}$ to $\mathbf{y}$ in the ambient space.

Authors of \cite{lcc09}, empirically show that when bases local to the input data $\mathbf{x}$ are preferred in the coding algorithm, the resultant sparse codes can improve the image classification accuracy; they conclude that locality is more essential than sparsity. Based on this idea, local coordinate coding (LCC) is proposed which uses the assumption that despite the features have a nonlinear structure in a high dimensional space, they lie on a manifold composed of locally linear patches. LCC achieved the state-of-the-art performance with a linear classifier in digit classification \cite{lcc09}. Nevertheless, due to it's high computational cost, it is not suitable for large-scale learning problems. In \cite{llc10}, a large-scale version of LCC named locality-constrained linear coding (LLC) is proposed. LLC guarantees locality by incorporating only the k-nearest bases in the coding process and minimizes the reconstruction term on the local linear patches of bases:
\begin{equation}
\label{lab:llceq}
\begin{split}
& \min_{\mathbf{c}_{knn}(\mathbf{x})} || \mathbf{x} - \mathbf{D}_{\mathbf{x}} \mathbf{c}_{knn}(\mathbf{x}) ||^2\text{,} \\
& \text{subject to: }  \mathbf{1}^\top \mathbf{c}(\mathbf{x}) = 1, \forall i
\end{split}
\end{equation}
where $ \mathbf{c}_{knn}( \mathbf{x})$ contains the non-zero coefficients of $\mathbf{c}(\mathbf{x})$ and the columns of matrix $\mathbf{D}_{\mathbf{x}}$ are the k-nearest bases to the data point $\mathbf{x}$. The dictionary are learned by k-means clustering algorithm. In recent years, many Locality-constrained coding methods have been proposed \cite{yu2010improved, zhang2011learning, huang2011salient, sac10, sac11, bo2009efficient} and applied successfully to image classification \cite{lcc09, llc10, sac10, sac11}, large scale object categorization \cite{lin2011large,xie2010large} and general object tracking \cite{liu2011robust}.

\subsection{Large Window Effect Drawback}
The assumption behind most of the locality constraint methods is that the datum and the bases, which are incorporated in its coding, lie on an almost linear local patch. Furthermore, recent researches on the coding problem \cite{lcc09,llc10,liu2011robust} show that sparse coding on many well-known datasets leads to locality preserving codes concomitantly. These researches reveal the fact that sparse coding usually works well when the bases that are incorporated in the coding of a datum lie close to a linear patch.

\begin{figure*}[t]
  \begin{minipage}{.5\textwidth}
    \centering
    \includegraphics[width=.4\textwidth]{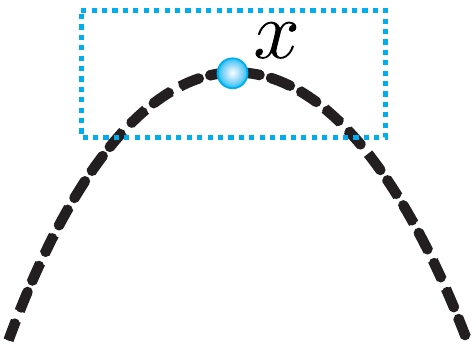}\quad
    \includegraphics[width=.4\textwidth]{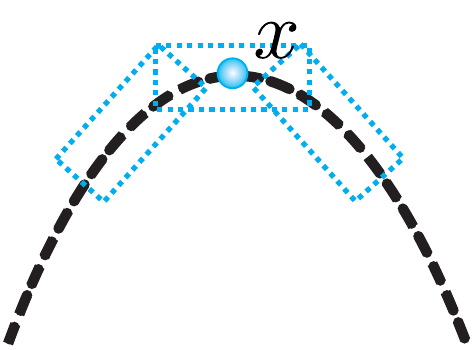}
    \subcaption{Local Coding vs. LSGC}
    \label{fig:lsgc}
  \end{minipage}
  \begin{minipage}{.5\textwidth}
    \centering
    \includegraphics[width=.4\textwidth]{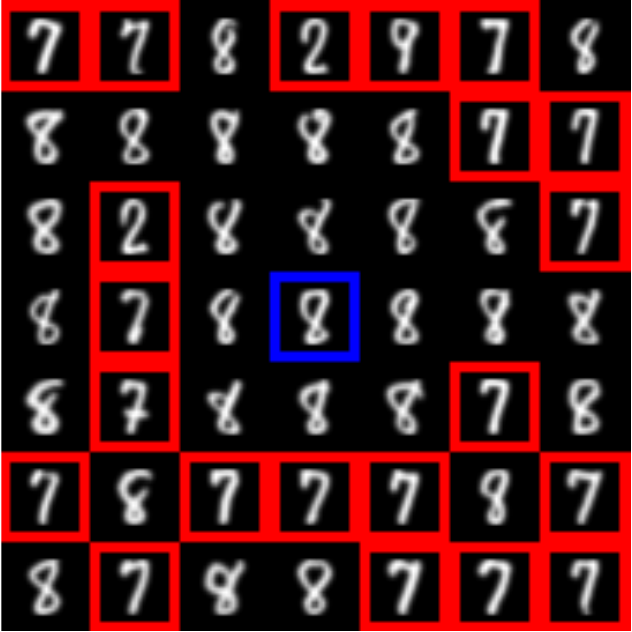}\quad
    \includegraphics[width=.4\textwidth]{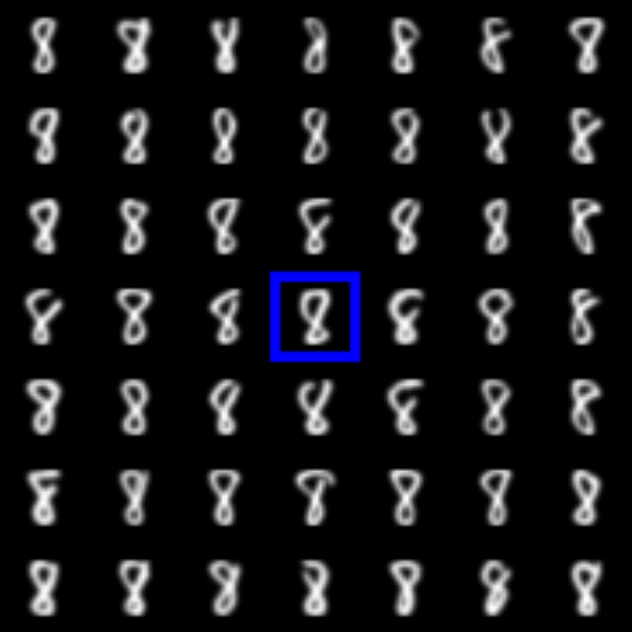}
    \subcaption{USPS}
    \label{fig:sub1}
  \end{minipage}\\[1em]
  
  \begin{minipage}{\textwidth}
    \centering
    \includegraphics[width=.3\textwidth]{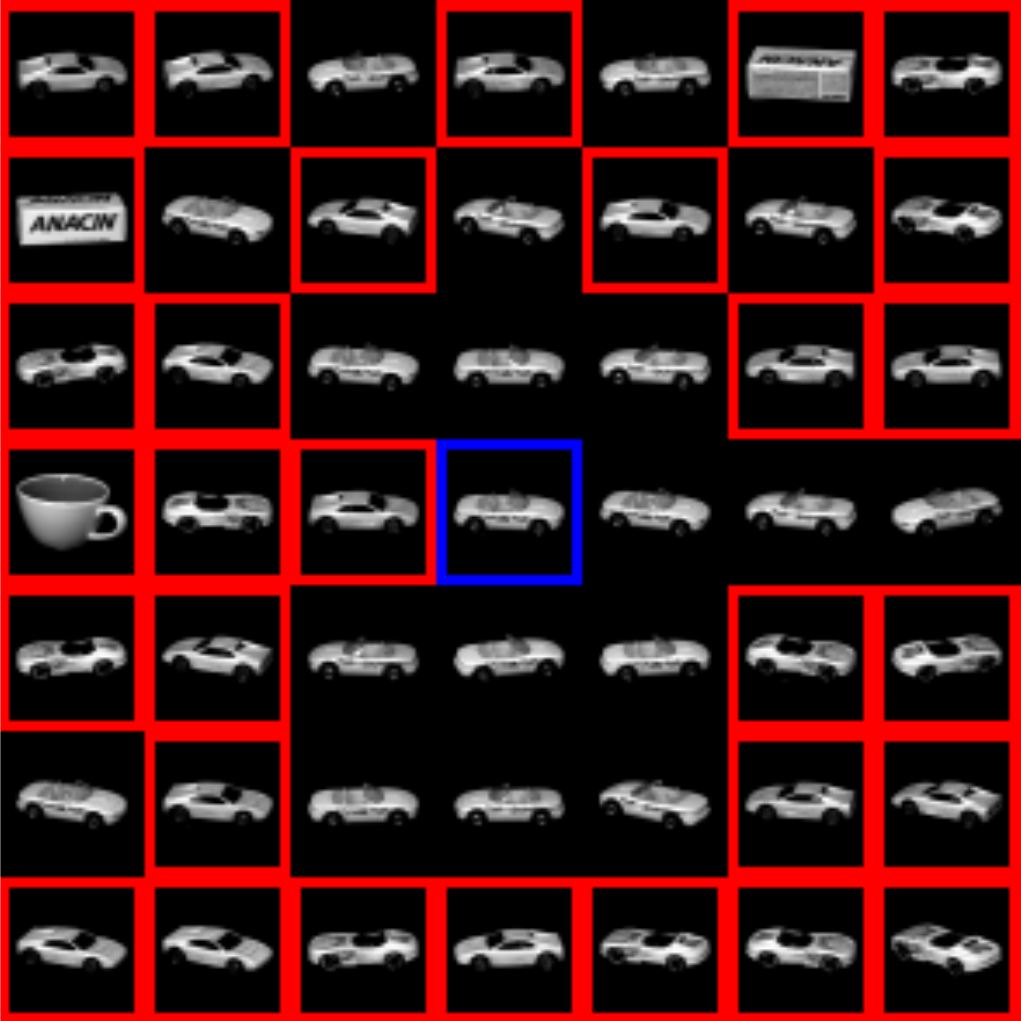}\quad
    \includegraphics[width=.3\textwidth]{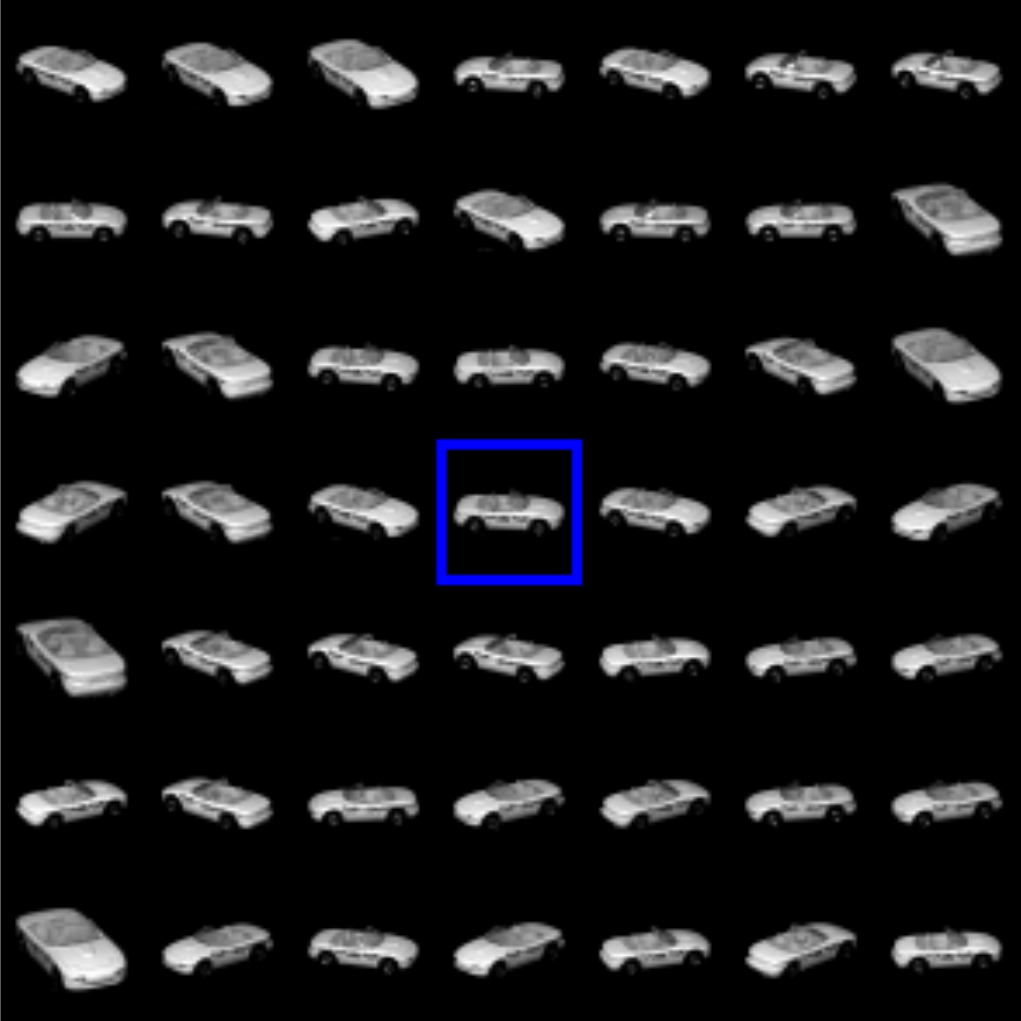}
    \subcaption{COIL20}
    \label{fig:sub2}
  \end{minipage}
  
  \caption{Comparing a Local coding method and LSGC: (a) The datum $\mathbf{x}$ is being coded with respect to the bases which lie on the nonlinear path. The window around the datum shows the set of local bases which are contributed in the coding of $\mathbf{x}$. A local coding method (left): the large window effect violates the assumption that local bases are lied on a one dimensional linear subspace, leads to an inaccurate code. LSGC (right): the coding coefficients are calculated accurately with respect to a set of small windows each containing a linear patch of the underlying structure. (b) and (c) compare the performance of the LLC and LSGC on USPS and COIL20 datasets respectively. Data points are sorted around the query image (blue) based on the value of the linear kernel function in the corresponding coding space. The linear kernel in the LLC coding space has high values for images from classes which are not the same as the class of the query image (red). In this experiment, difficult query images are selected to clarify the effect of choosing small window sizes. Note that parameter of the LLC and LSGC are tuned to have their best accuracy in the image classification task.}
\end{figure*}

However, in some situations this assumption may not be completely true. As an example, in Figure \ref{fig:lsgc} where the datum $\mathbf{x}$ is coded with respect to the bases which lie on the nonlinear path. The window around the datum shows the set of local bases which are used in its code. It is clear that as the window size grows, the error of approximating the local bases structure by a linear one increases and the coding methods based on this simplifying assumption fail. While we explained the situation in a rather simple example, our experiments show that these coding methods have a tendency to use bases outside the local linear patches to code the data in real world datasets. This phenomena may be a result of one or some of the following reasons:
\begin{itemize}
\item[1)]{\bf Lack of labeled points:} representer theorem \cite{bishop2006pattern} states that the SVM labeling function can be formulated as:
\begin{equation}
f = \sum_i \beta_i \mathcal{K}(\mathbf{.}, \mathbf{x_i})
\end{equation}
where $\mathcal{K}(\mathbf{x},\mathbf{y}) = \mathbf{\Phi}(\mathbf{x})^\top .  \mathbf{\Phi}(\mathbf{y})$ is the defined linear kernel function in the coding space, $\beta_i$s are Lagrange coefficients, and $\mathbf{x_i}$s are the support vectors. When the number of labeled points is limited, there are many data points for which the set of bases lie in their local linear patch has no intersection with those in the linear patches around the support vectors. For these points we have $\mathbf{\forall} i, \mathcal{K}(\mathbf{x}, \mathbf{x_i}) \approx 0$ and SVM fails to predict the label of $\mathbf{x}$. As a result, in the cross validation phase the parameters of the coding method are chosen in a way that the number of bases which contribute in the code of each datum increases and bases are selected far from the local linear patch around the datum.

\item[2)] {\bf Smoothness of labeling function:} many learning algorithms are based on the assumption that the labeling function changes smoothly with respect to the underlying distribution of the data \cite{zhu2005semi,belkin06}.  Given that the smoothness assumption holds for a dataset, it is more likely that two data points that do not share their linear local neighborhoods be still close enough to be considered in the same context. In this case, for the kernel function to be capable of capturing the similarities among these data points, it must have reasonable non-zero values within a more global non-linear patch. Consequently, the encoder tries to improve the performance of the learning algorithm by using bases that do not lie on the linear local patch around the data point being coded.

\item[3)] {\bf Distinctive neighborhood structure:} since distribution of bases varies in different regions of the feature space, the neighborhood structure of each data point should be distinctive in each region. However, both SAC and LLC methods use a globally fixed number of neighborhood bases to achieve a good performance on average.  Considering a fixed number of bases around each data point and the variations in the distribution density of bases, there are some regions with few selected bases and regions in which large number of bases outside the linear local patch of a feature are selected.


\item[4)] {\bf Insufficient number of bases:} due to computational complexity and memory limitations, number of bases is usually limited to some thousands. Therefore, there will be some regions in the feature space in which the set of selected bases in the linear local patch around a data point is sparse, leading to an unstable and inaccurate coding. For alleviating the problem, again it is reasonable that the coding method use bases out of the local patch in these sparse regions.

\end{itemize}

While growing the window size increases the performance of the algorithm by offering a more global kernel, the quality of the coding method degrades since the selected bases do not lie close to a linear patch and the primal assumption behind these local encoders is not tenable anymore. 
\subsection{Paper Contributions}
In this paper, to solve the aforementioned problem, we propose a method dubbed Local Similarities Global Coding (LSGC), in which the coding coefficients are calculated accurately with respect to a set of small windows each containing a linear patch of the underlying bases structure.  A random walker connects the coding information in each window together and computes the final code for each data point. 
To be more precise, first we calculate the coding vector of each basis with respect to the other bases which somehow captures the underlying local structure of bases. 
Then, a weighted graph is constructed whose nodes are  $\mathbf{S} = \{\mathbf{x}, \mathbf{b_1}, \dots, \mathbf{b}_K \}$ and the weight $\mathbf{R}(\mathbf{x}, \mathbf{b_i})$ is deduced from the coding of instance $\mathbf{x}$ corresponding to the basis $\mathbf{b_i}$. In a similar way, $\mathbf{R}(\mathbf{b_i}, \mathbf{b_j})$ is deduced from coding of basis $\mathbf{b_i}$ corresponding to the basis $\mathbf{b_j}$. 
Finally, the coding for $\mathbf{x}$ can be obtained naturally from a random walk on the constructed graph. Though this encoding appears to be similar to Markov random field for classification task \cite{jaakkola2002partially}, the LSGC differs basically, because LSGC runs on the graph whose nodes are learned visual bases. The process is depicted in Figure \ref{fig:lsgc}. By using small windows to calculate the coding vectors, the coding algorithm is free to use bases outside the local linear patch around each data point while the estimated codes remain accurate. 

Theoretically, we show that the linear kernel function in the LSGC coding space approximates the diffusion kernel \cite{lafon05} that is used  repeatedly in manifold learning applications. It is shown that the approximation error converges to zero with the rate of $\mathcal{O}(K^{-1})$ where $K$ is the number of bases. Compared to the diffusion kernel, LSGC coding method is inductive and fast which makes it appropriate for large scale settings. Furthermore, by mapping the input data to the coding space and then applying a fast linear method the training and testing are accelerated compared to the training and testing of a kernel machine that uses diffusion kernel.

A preliminary conference version of this paper appears in \cite{me2013}. We extend our work in the following three aspects: 1) Compared to \cite{me2013}, here we present a theoretical justification for the proposed encoding algorithm, which reveals the key components of its success as a coding scheme.  LSGC encoder based on LLC and sparse coding besides the SAC encoder which appeared in \cite{me2013}. 2) To show that the proposed method can be used as a framework in coding applications, we introduce and study different versions of LSGC encoder based on LLC and sparse coding besides the SAC encoder which appeared in \cite{me2013}. 3) To validate the effectiveness of LSGC encoder we expand our experiments on several datasets in two ways. First, the behaviors of the proposed methods are studied under a vast variety of new settings. Second, we add clustering and regression problems as two new learning tasks to support the main contribution of the paper more thoroughly.

The rest of paper is organized as follows. In Section \ref{lab:LSGC}, assuming that the relations $\mathbf{R}$ are defined for each coding method, we propose the LSGC encoding algorithm. Section \ref{lab:relations} discusses the ways which are used to define the $\mathbf{R}$ matrix for each encoding method. We present our experimental evaluations of the encoding algorithm on three learning task in Section \ref{lab:experiments} and compare the proposed method to several other encoding approaches. Finally, in Section \ref{lab:conclusions} we conclude the paper and provide the future works.

\section{Proposed Method}
\subsection{Local Similarities Global Coding}
\label{lab:LSGC}
Let $\mathbb{G}(\mathbf{V}, \mathbf{R} )$ be a graph whose nodes $\mathbf{V}$ are the set of bases, and $\mathbf{R}$ is a positive symmetric matrix (we let $\mathbf{R} \leftarrow \big(\mathbf{R} + \mathbf{R}^\top \big) / 2$ if $\mathbf{R}$ is not symmetric ) that captures the pairwise relations between bases and $\mathbf{l(x)}$ is the coding vector of $\mathbf{x}$ with positive elements which are computed by the corresponding coding algorithm. The computation of $\mathbf{R}$ and $\mathbf{l}$ depends on the coding algorithm being used in our framework and will be discussed shortly in this section.

As a start point towards the calculation of our coding method, first we symmetrically normalize the matrix $\mathbf{R}$ as follows:
\begin{equation}
\tilde{\mathbf{P}} = \mathbf{D}^{-1/2} \mathbf{R} \mathbf{D}^{-1/2} 
\end{equation}
in which $\mathbf{D}$ is a diagonal matrix and each of its elements $\mathbf{D}_{ii}$ denotes the sum of the degrees of the edges connecting the $i$'th basis to others. It can be easily shown that by this normalization the (i,j)'th element of the matrix $\tilde{\mathbf{P}}$ that shows dependency between basis $\mathbf{b_i}$ and $\mathbf{b_j}$ can be calculated as:
\begin{equation} \label{equ_normalization}
\tilde{p}^1(\mathbf{b_i}, \mathbf{b_j}) = \sqrt{\frac{d(\mathbf{b_i})}{d(\mathbf{b_j})}} p^1(\mathbf{b_i}, \mathbf{b_j}) = \frac{\mathbf{R}(\mathbf{b_i}, \mathbf{b_j})}{\sqrt{d(\mathbf{b_i}) d(\mathbf{b_j})}}
\end{equation}
where $d(.)$ calculates the degree of the nodes and $p^1(\mathbf{b_i}, \mathbf{b_j})$ is the probability of going from $\mathbf{b_i}$ to $\mathbf{b_j}$ in a one-step random walk transition: 
\begin{equation} \label{equ_p}
p^1(\mathbf{b_i}, \mathbf{b_j}) = \frac{\mathbf{R}(\mathbf{b_i}, \mathbf{b_j})}{d(\mathbf{b_i})}.
\end{equation}

Although $\tilde{p}^1$ is calculated based on the one-step transition probabilities, it can be shown that, like $p^1$, its t-step version which is denoted by $\tilde{p}^t$ can be computed by the Chapman-Kolmogorov equation.

While conventional coding algorithms can be viewed as a one-step random walk in our framework, the proposed method utilizes t-step walks to capture the nonlinear dependencies among the bases as follows:
\begin{equation}
\label{eq_coding_coeff}
\mathbf{c}_{(i)}^t(\mathbf{x}) = \sum_{k = 1}^{K} \tilde{p}^1(\mathbf{x}, \mathbf{b_k}) \tilde{p}^{t-1}(\mathbf{b_k}, \mathbf{b_i})
\end{equation}
where $\mathbf{c}_{(i)}^t(\mathbf{x})$ is the LSGC coding coefficient of $\mathbf{x}$ with respect to the basis $\mathbf{b_i}$. This coding method can be formulated as a simple matrix-vector multiplication as follows:
\begin{equation}
\label{eq:lsgc}
\mathbf{c}^t(\mathbf{x}) = \mathbf{\tilde{P}}^{t-1} \mathbf{\tilde{l}}(\mathbf{x})
\end{equation}
in which elements of vector $\mathbf{\tilde{l}}(\mathbf{x})$ are $\tilde{p}^1(\mathbf{x}, \mathbf{.})$ that are computed by normalizing the original local coding vector $\mathbf{l}(\mathbf{x})$ similar to equation (\ref{equ_normalization}):
\begin{equation}
\mathbf{\tilde{l}}_{(i)}(\mathbf{x}) = \frac{\mathbf{l}_{(i)}(\mathbf{x})}{\sqrt{d(\mathbf{x}) \big(d(\mathbf{b_i}) + \mathbf{l}_{(i)} (\mathbf{x}) \big)}}
\end{equation}  
where $d(\mathbf{b_i}) + \mathbf{l}_{(i)} (\mathbf{x})$ is the degree of node $\mathbf{b_i}$ after adding the datum $\mathbf{x}$ into the graph $\mathbb{G}$. Since $\tilde{\mathbf{P}}^{t-1}$ can be precomputed with the cost of $\mathcal{O}(K^3)$, the complexity of coding the data points is $\mathcal{O}(nKz + K^3)$ in which $n$ is the number of images, $K$ is the number of visual bases and $z$ is the number of non-zero elements in $\tilde{\mathbf{l}}(\mathbf{x})$ which is small due to the locality or sparsity constraint. In many learning tasks, the number of images can be in the order of thousands. However, the number of bases are constant and relatively small (normally around hundreds of units). This make our algorithm of practical interest when the number of images are relatively large. This computational cost is reasonable compared to the other algorithms reviewed in previous sections, that run with the cost of $\mathcal{O}(nK)$.

While for computational efficiency, in practice, a linear learning algorithm is used explicitly after mapping the points to the coding space, studying the kernel behind a coding algorithm can help assess the properties of the coding scheme. Given that a linear learning algorithm is used in the coding space, the problem can be viewed as a nonlinear one based on a kernel corresponding to the coding algorithm.

\begin{lemma}\label{MahLemma}
The LSGC coding vector is related to diffusion kernel of order $t$ with the following equation: 
\begin{equation}
\tilde{p}^{2t} (\mathbf{x}, \mathbf{y}) = \mathcal{K}^{2t} (\mathbf{x}, \mathbf{y}) + \mathbf{r}^{2t} (\mathbf{x}, \mathbf{y})
\end{equation}
where $\mathcal{K}^{2t} (\mathbf{x}, \mathbf{y})$ is the LSGC kernel function and the residual term $\mathbf{r}^{2t} (\mathbf{x}, \mathbf{y})$ is the value of $\tilde{p}^{2t}$ for the paths with $2t$-steps in which $\mathbf{x}$ is visited at least two times or $\mathbf{y}$ is visited at least two times.\hfill\(\Box\)
\end{lemma}
The lemma can be easily validated algebraically by expanding $\mathcal{K}^{2t} (\mathbf{x}, \mathbf{y}) = \mathbf{c}^t (\mathbf{x})^\top . \mathbf{c}^t (\mathbf{x})$, using the  Chapman-Kolmogorov equation and considering the fact that the paths which contribute in $\tilde{p}^{t-1}(\mathbf{b_i,b_j})$ in equation (\ref{eq_coding_coeff}) do not visit $\mathbf{x}$ or $\mathbf{y}$ since they are not included in the graph $\mathbb{G}$. 

As is seen in Lemma \ref{MahLemma}, if the residual term is small enough with respect to $\tilde{p}^{2t} (\mathbf{x}, \mathbf{y})$, the kernel behind the proposed method is an approximation of the well-known diffusion kernel $\tilde{p}^{2t} (\mathbf{x}, \mathbf{y})$ \cite{lafon05}, which calculates similarities between the graph nodes by considering all the paths connecting them with a specific step length. Since close points over the graph are connected with several paths to each other, this kernel assigns higher values to close points over the underlying structure. It is worth noting that the diffusion kernel has successfully been applied to several applications in manifold learning problems \cite{lerman07, singer09}. 

\begin{theorem}
$\frac{\mathbf{r}^{2t} (\mathbf{x}, \mathbf{y})}{\tilde{p}^{2t} (\mathbf{x}, \mathbf{y})}$ converges to zero at the rate of $\mathcal{O}(K^{-1})$ where $K$ is the number of visual bases in the graph.
\end{theorem}
\subsection{Computing the Relation Matrix $\mathbf{R}$}
\label{lab:relations}
Until now, we assume that the relation matrix is given. Since any encoding algorithm used in the framework has its own nature and properties, a specific matrix $\mathbf{R}$  needs to be computed for each coding algorithms.  In this subsection, methods for computing $\mathbf{R}$ are proposed for SAC \cite{sac10}, sparse coding \cite{scspm09} and LLC \cite{llc10}. 
The details for constructing the adjacency matrices for each of these encoding algorithms follows:

\subsubsection{\bf Soft Assignment Coding}
$\mathbf{R}$ can easily be computed for SAC from the original formulation of the algorithm:
\begin{equation}
\mathbf{R}(\mathbf{b_i}, \mathbf{b_j}) = \exp \Big(- \frac{|| \mathbf{b_i} - \mathbf{b_j} || ^ 2}{2 \sigma^2} \Big)
\end{equation}
since $\mathbf{R}$ is a non-negative matrix, it can be used directly in the LSGC coding method. Similarly $\mathbf{l}(\mathbf{x})$ is computed with the Gaussian kernel function.

\subsubsection{\bf Sparse coding}
the pairwise relations between bases for this coding method is based on the contribution of the bases in the sparse codes of each other. For this purpose, we exclude the basis that is being coded from the dictionary and compute its sparse code with the original formulation of the algorithm:
\begin{equation}
\begin{split}
& \min_{\mathbf{R^\prime(b_i, .)}} || \mathbf{b_i} - \mathbf{D}_\mathbf{b_i} \text{ } \mathbf{R^\prime(b_i, .)} ^\top ||_2 ^ 2 + \lambda || \mathbf{R^\prime(b_i, .)} ||_1  \\
&\text{subject to: } \mathbf{R^\prime}(\mathbf{b_i}, \mathbf{b_i}) = 0
\end{split}
\label{eq:sparse_R_eq}
\end{equation}
where $\mathbf{D}_\mathbf{b_i}$ is the dictionary without the basis $\mathbf{b_i}$. The matrix $\mathbf{R^\prime}$ which is learned in this way may have negative values. By duplicating the number of bases and letting $\mathbf{D_\mathbf{b_i}} \leftarrow [ \mathbf{D}_\mathbf{b_i},-\mathbf{D}_\mathbf{b_i}]$, the sparse codes will have the length of $2K$. However, the result of the new optimization problem can be determined by the solution of equation (\ref{eq:sparse_R_eq}): sparse codes for the first $K$ bases will be $\mathbf{R}(\mathbf{b_i}, \mathbf{.}) \leftarrow [\mathbf{R^\prime}_p(\mathbf{b_i}, \mathbf{.}), \mathbf{R^\prime}_n(\mathbf{b_i}, \mathbf{.})]$ and the sparse codes for the remaining $K$ bases ( $K < i <= 2K$ ) are  $\mathbf{R}(\mathbf{b_i}, \mathbf{.}) \leftarrow [\mathbf{R^\prime}_n(\mathbf{b_{i - K}}, \mathbf{.}),\mathbf{R^\prime}_p(\mathbf{b_{i - K}}, \mathbf{.})]$ where $\mathbf{R^\prime}_n(\mathbf{b_i}, \mathbf{.}) = \max(-\mathbf{R^\prime}(\mathbf{b_i}, \mathbf{.}), \mathbf{0})$ and $\mathbf{R^\prime}_p(\mathbf{b_i}, \mathbf{.}) = \max(\mathbf{R^\prime}(\mathbf{b_i}, \mathbf{.}), \mathbf{0})$. Therefore the non-negativity constraint is trivially satisfied. The positive relation matrix of these $2K$ bases can be written as:
\begin{equation}
\mathbf{R} \leftarrow 
\left[ {\begin{array}{*{20}c}
\mathbf{R^\prime}_p & \mathbf{R^\prime}_n  \\
\mathbf{R^\prime}_n & \mathbf{R^\prime}_p  \\
\end{array} } \right].
\end{equation}
In the same way, we let $\mathbf{l}(\mathbf{x}) \leftarrow [\mathbf{l}_p(\mathbf{x}), \mathbf{l}_n(\mathbf{x})]$ and compute the LSGC based on the sparse coding algorithm. Finally, after computing the LSGC codings, positive and negative parts of the codes are merged together in order to have a code with length $K$.

\subsubsection{\bf Locality-constrained Linear Coding} 
Based on the LLC encoding method in equation (\ref{lab:llceq}), the non-zero elements of relation matrix can be computed as:
\begin{equation}
\min_{\mathbf{R_{knn}(b_i, .)}} || \mathbf{b_i} - \mathbf{R_{knn}(b_i, .) \mathbf{D_{\mathbf{b_i}}} } ||_2 ^ 2.
\end{equation}
where columns of $\mathbf{D_{\mathbf{b_i}}}$ are the $k$-nn bases of $\mathbf{b_i}$ and elements of $\mathbf{R_{knn}(b_i, .)}$ are the non-zero relations between $b_i$ and its $k$-nn bases. Due to the locality constraint of the LLC method the situation here is a little different from sparse coding and the non-negativity of the relation matrix can not be truly satisfied by duplicating the bases. By adding $-\mathbf{D_{\mathbf{b_i}}}$ to the local dictionary, the new bases are not in the locality of basis that is being coded and the basic assumption with the LLC method is not satisfied.  However, the amplitude of coefficients in the LLC code naturally captures the bases relations in the feature space. As a consequence, as is suggested previously \cite{l1graph13}, we use the absolute value of the coefficients as weights among bases: $\mathbf{R} \leftarrow  |\mathbf{R} |$. In the same way, we let $\mathbf{l}(\mathbf{x}) \leftarrow | \mathbf{l}(\mathbf{x}) |$ to satisfy the positivity constraint.

\section{Experiments}
\label{lab:experiments}
In this section, we evaluate the performance of the proposed method with different settings on five real world and one artificial datasets. The detailed characteristics of each real world dataset is shown in Table \ref{tab:dataset}. The datasets belong to three main categories of objects, hand written digits, and hand written letters. Size of datasets vary from 1440 to 70000 instances and the number of attributes vary from 16 to 1024. In sparse coding algorithm, features are normalized to have unit norm for deriving semantically reasonable coding vectors. It is worth to mention that, the processing time of our method is very close to the original encoding algorithm. This is due to the fact that the cost of computing the matrix $\tilde{P}^{t-1}$ and the matrix vector multiplication in the equation (\ref{eq:lsgc}) is very low compared to the computational cost of the corresponding coding method and takes less than \%$5$ of the CPU time on the average. Consequently, the processing times are not reported in the tables. Detailed settings of algorithms are described for each experiment accordingly.

\begin{table}[h]
\centering
\caption{Characteristics of real world datasets.}
\begin{tabular}{| c | c | c | c | c |}
\hline
\bf{Dataset Name} & \bf{\#Instances} & \bf{\#Attributes} & \bf{\#Classes} \\
\hline
COIL20 & 1440 & 1024 & 20\\
\hline
Digit \cite{uci10} & 5620 & 64 & 10\\
\hline
USPS & 9298 & 256 & 10\\
\hline
Letter \cite{uci10} & 20000 & 16 & 26\\
\hline
MNIST & 70000 & 784 & 10\\
\hline
\end{tabular}
\label{tab:dataset}
\end{table}


We used the implementation of LLC and Regularized sparse coding which have been made public by the authors. LIBSVM \cite{libsvm11} package is used for regression and LIBLINEAR \cite{liblinear08} is used for the linear classification task.
\subsection{Regression}
\begin{figure}[t]
\centering
\includegraphics[width=\columnwidth]{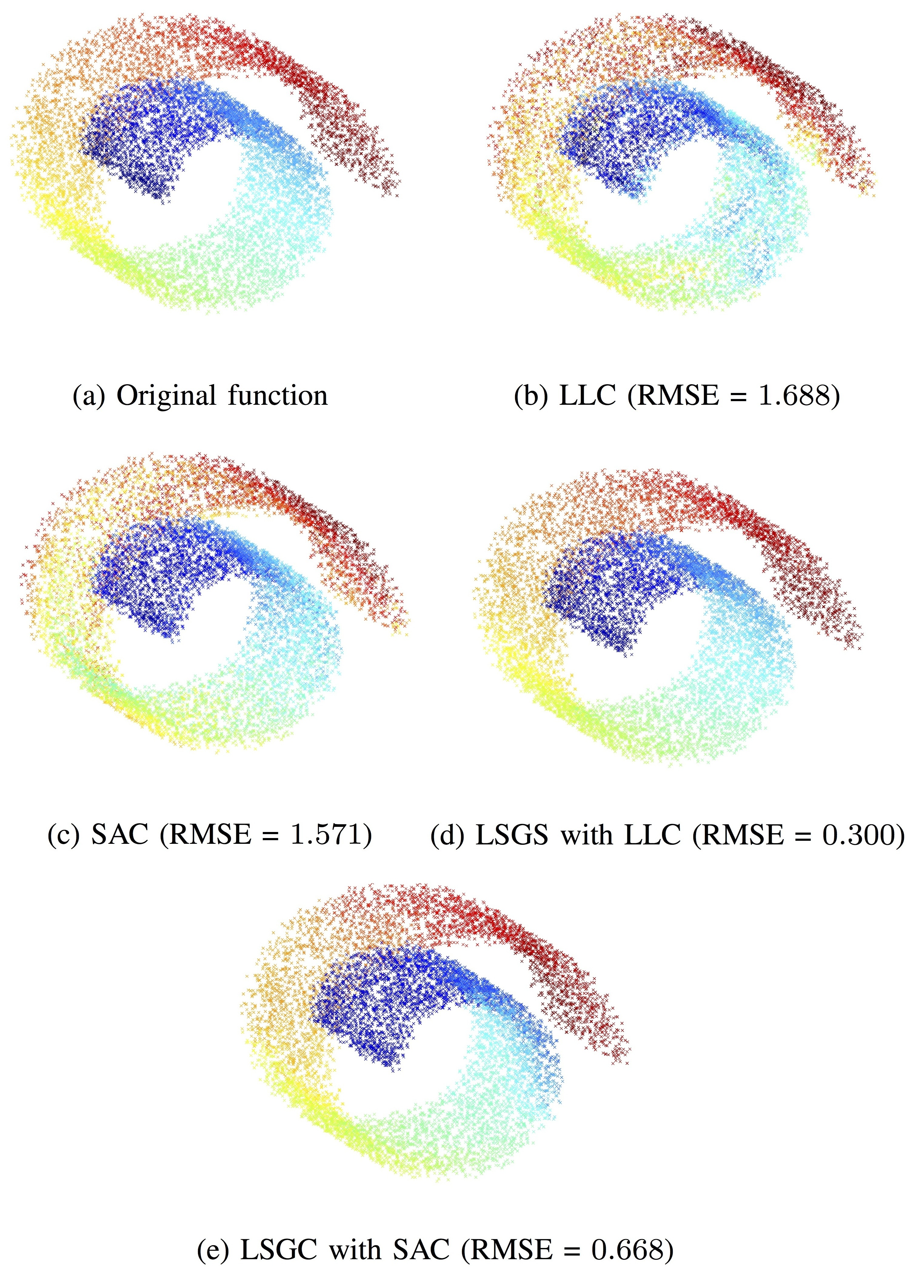}
\caption{Nonlinear regression results on the Spiral dataset. The goal is to learn the nonlinear function in figure (a). In each figure the colors indicate the values of function in each data point. Root Mean Square Error (RMSE) of ridge regression method is reported for each method.}
\label{fig:spiral}
\end{figure}

In this part, the primary goal is to learn a nonlinear function defined on a spiral dataset. For this experiment, $10,000$ points are sampled from the well-known spiral shaped manifold as shown in Figure \ref{fig:spiral}a. For each experiment, $512$ bases are learned and $100$ data points are used for training. The performance of different algorithms are compared in the regression task using the ridge regression algorithm in the coding space. The results are shown in Figure \ref{fig:spiral}. Each image shows the results of regression with different coding methods on the corresponding function. The average Root Mean Squared Error (RMSE) for $20$ independent runs are also reported for each method. In all methods, the parameters are tuned to achieve to their best result. The sparse coding method is intentionally excluded from this experiment, since, as is reported in \cite{lcc09}, sparse coding fails to capture the nonlinear structure of the spiral and has a poor performance on this toy dataset. 

Since LLC method does not employ the smoothness assumption, it is not able to predict the values of the function in points that are distant from the labeled points.  
However, LSGC has a much better performance and predicts the labels of unlabeled points which are far from the labeled points accurately.  On the other hand, although result of SAC is smooth, in some regions is shifted from the true values. The smoothness of the predicted function is due to the smooth nature of the Gaussian kernel, but considering that SAC uses Euclidean distance, the value of function in some parts is highly affected by others that have a small Euclidean distance but a large manifold distance. 

\begin{figure}[h]
        \centering
        \begin{subfigure}[b]{0.45\columnwidth}
                \centering
                \includegraphics[width=\textwidth]{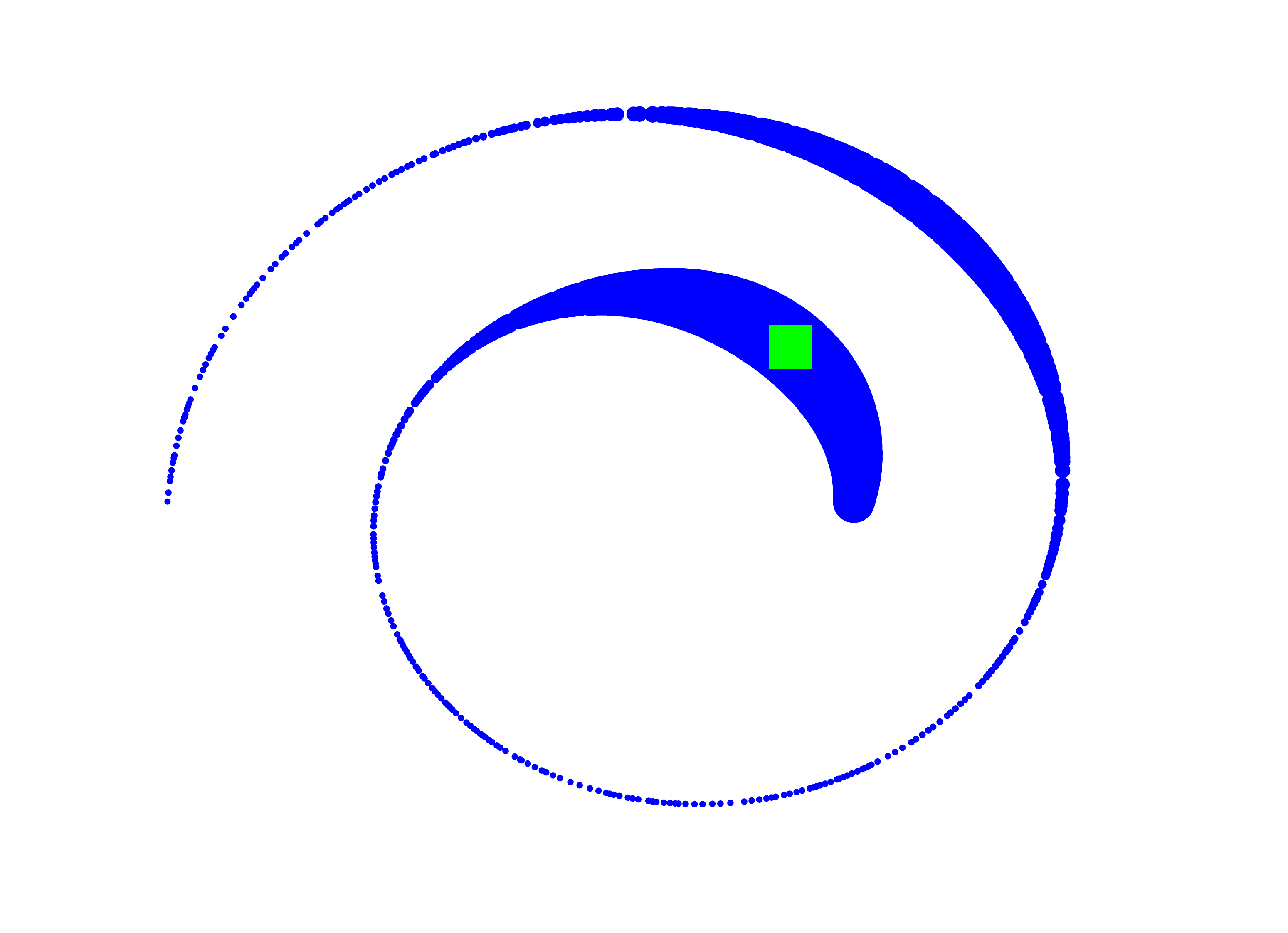}
                \caption{SAC}
                \label{fig:contributions_SAC}
        \end{subfigure}%
        ~
        \begin{subfigure}[b]{0.45\columnwidth}
                \centering
                \includegraphics[width=\textwidth]{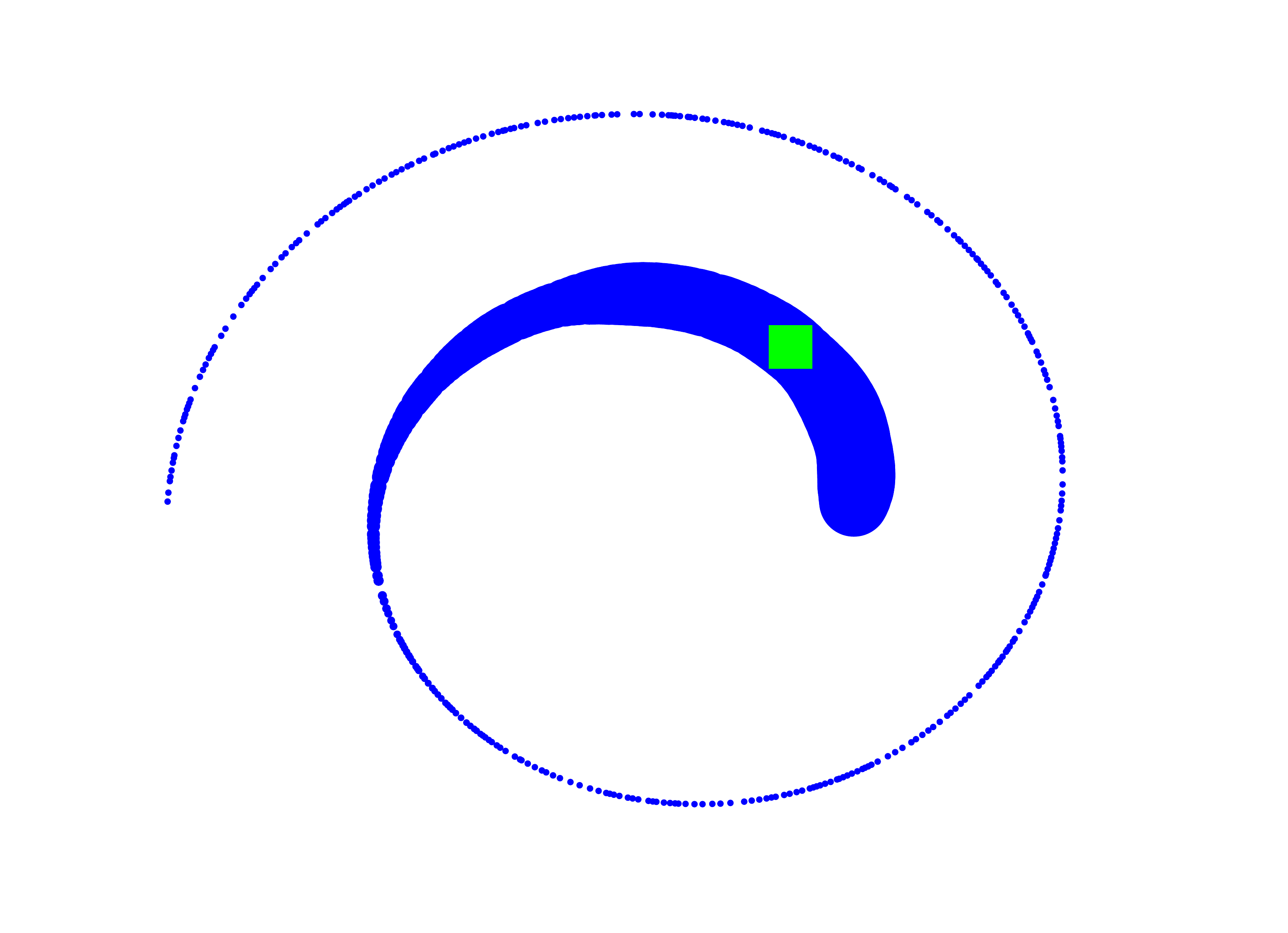}
                \caption{LSGC with SAC}
                \label{fig:contributions_SAC2}
        \end{subfigure}
        
        \begin{subfigure}[b]{0.45\columnwidth}
                \centering
                \includegraphics[width=\textwidth]{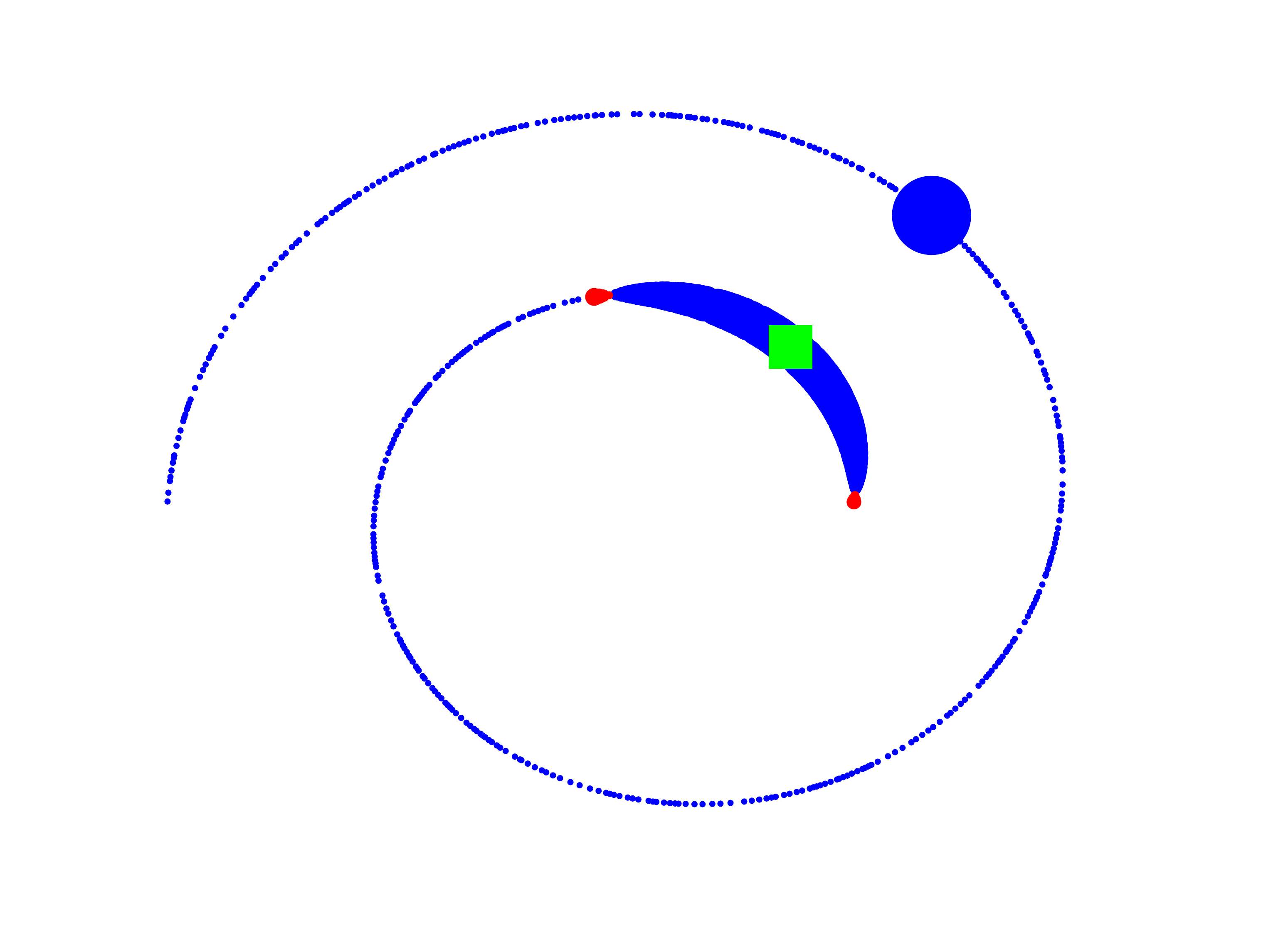}
                \caption{LLC}
                \label{fig:contributions_LLC}
        \end{subfigure}%
        ~   
        \begin{subfigure}[b]{0.45\columnwidth}
                \centering
                \includegraphics[width=\textwidth]{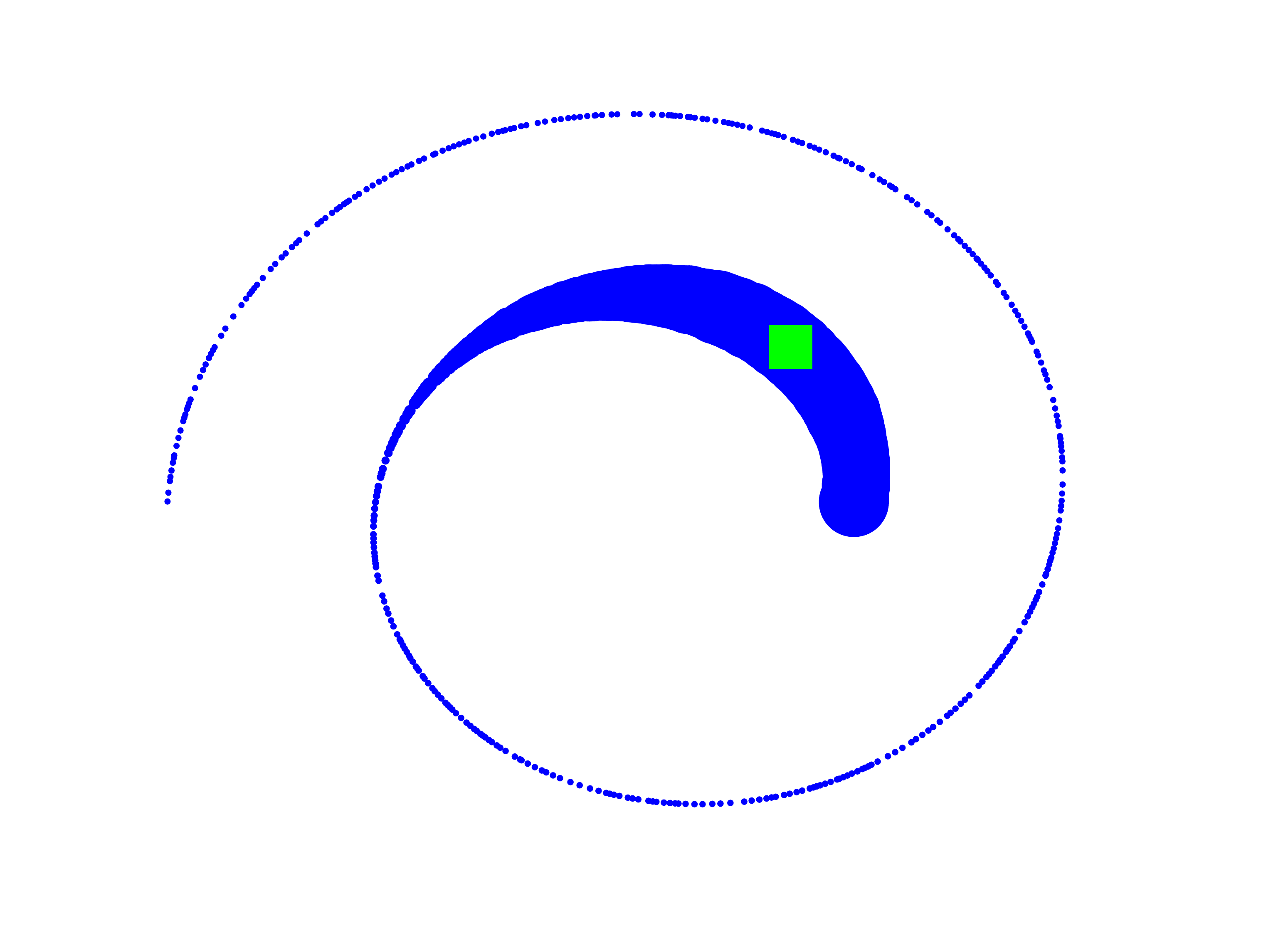}
                \caption{LSGC with LLC}
                \label{fig:contributions_LLC2}
        \end{subfigure}
        \caption{Contribution of bases for a sample data point. Filled circles show the bases position and the radius of bases reflects the amplitude of the corresponding coding coefficients in coding of query data point which is depicted by a square in the figure. The blue and red colors show positive and negative coefficient values respectively.}
        \label{fig:contributions}
\end{figure}

To take a closer look at how each method assigns coding coefficients to a data point, we depicted the contribution of each bases for a sample data point in Figure \ref{fig:contributions}, after eliminating one of the dimensions for better visibility. It is seen that although SAC assigns codes in a smooth manner with respect to the Euclidean distance of the bases from the data point, the same cannot be said if the manifold distance is considered. In fact, bases with a large manifold distance from the sample point have contributions that may mislead the regressor and cause a shift from the true values as discussed above. Similarly, LLC assigns meaningful codes for bases which are near to the data point, while for the bases outside its linear local patch, the performance of the algorithm degrades. However, LSGC (Figure \ref{fig:contributions_SAC2} and \ref{fig:contributions_LLC2}) leads to a more global coding vector in which the contribution of the bases in the coding vector reduces with respect to their manifold distance from the data point. For LLC and SAC, a large window size results in degraded classification accuracy since coding coefficients are assigned with respect to the Euclidean distance to the bases. This phenomena restricts LLC and SAC to use bases that are far from the linear local patch of the data point.

\subsection{Image Classification}
In this section we aim to compare the classification performance of LSGC to the original representation (OR) and three types of coding methods: 1) Similarity based: Soft Assignment Coding (SAC) \cite{sac10}, 2) Locality based: Local Coordinate Coding (LCC) \cite{lcc09}, Locality-constrained Linear Coding (LLC) \cite{llc10}, and 3) Sparsity based: Sparse Coding (SC) \cite{scspm09}, Regularized Sparse Coding (RSC) \cite{zheng2011graph}. For each method, we train a SVM with the linear kernel in the coding space for each class and use one against all method to evaluate the performance of different algorithms.


In all methods the parameters are set by 5 fold cross validation. A wide range of test values are selected to ensure that the proper value for each method is tested in the cross validation. The test values for sparse regularization parameter $\lambda$ are $\{0.005, 0.01, 0.05, 0.08, 0.1, 0.2, 0.3, 0.4\}$. In LLC we test $\{ 3, 5, 7, 10, 30, 50, 100, 200, 400\}$ values for $k$-nn. We test $\{ 0.05, 0.1, 0.2, 0.3, 0.4, 0.5, 0.6, 0.7, 0.8, 0.9, 1\} \times \bar{\sigma}$ to find best bandwidth parameter $\sigma$ in SAC. The scaling factor $\bar{\sigma}$ is the mean of the standard deviation of the data points and is used since the proper value of $\sigma$ depends on the variance of the data points.
The test parameters used for RSC are as those reported in \cite{zheng2011graph}.  As we explore later, LSGC results are not very sensitive to the parameter of its base method, if the base method coding vectors are local enough. Then the globalization of the final coding vector can be controlled by the step size parameter $t$. Thus, we test limited values of $k = \{ 3, 5, 7 \}$, $\sigma = \{ 0.1, 0.3, 0.6 \}$ and $\lambda = \{ 0.1, 0.2, 0.4\}$ for the base methods and the step size parameter $t$ is selected from $\{ 1, 2 ,3 ,5, 8, 11, 15\}$. For each algorithm, the best performance over the dictionary size of $\{128, 256, 512, 1024 \}$  is reported. In Tables \ref{c_COIL20}-\ref{c_USPS} the average performance of the algorithms over $20$ independent run is shown. Generally, LSGC with $t > 1$ improves the accuracy of its base methods by propagating the coefficients. The improvement is more tangible especially when the number of the labeled data point is limited. However, in Letter dataset the LSGC with SAC and LLC fails to improve the performance of the base methods. The reason is that, roughly speaking, Letter is a low dimensional dataset, thus bases could not reside on a much lower dimensional structure.

\begin{figure}[h]
        \centering
        \begin{subfigure}[b]{1\columnwidth}
                \centering
                \includegraphics[width=.9\textwidth]{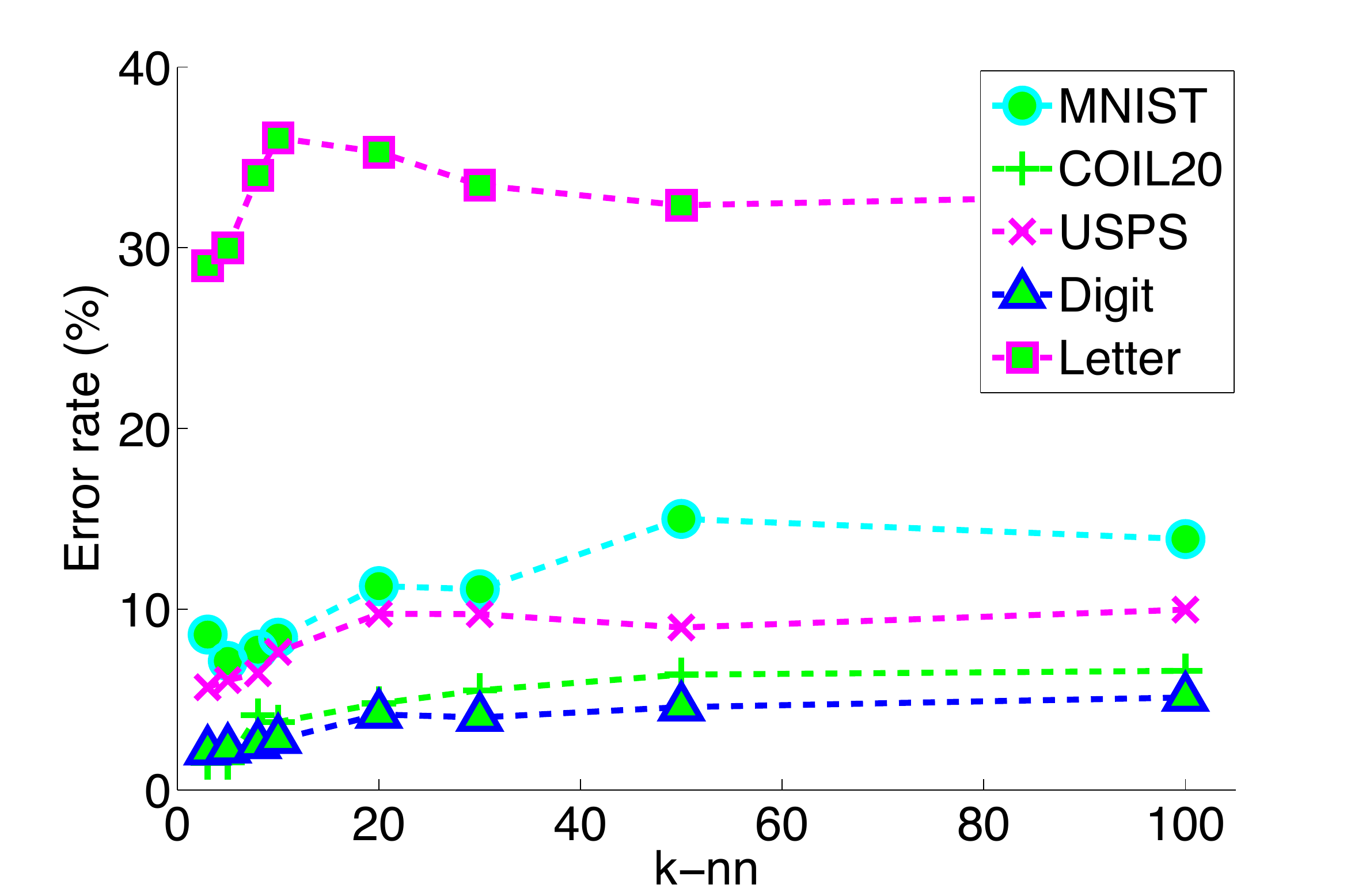}
                \caption{LSGC + LLC method (error rate (\%) vs. $k$-nn)}
                \label{fig:knnEffect}
        \end{subfigure}
        
        \begin{subfigure}[b]{1\columnwidth}
                \centering
                \includegraphics[width=.9\textwidth]{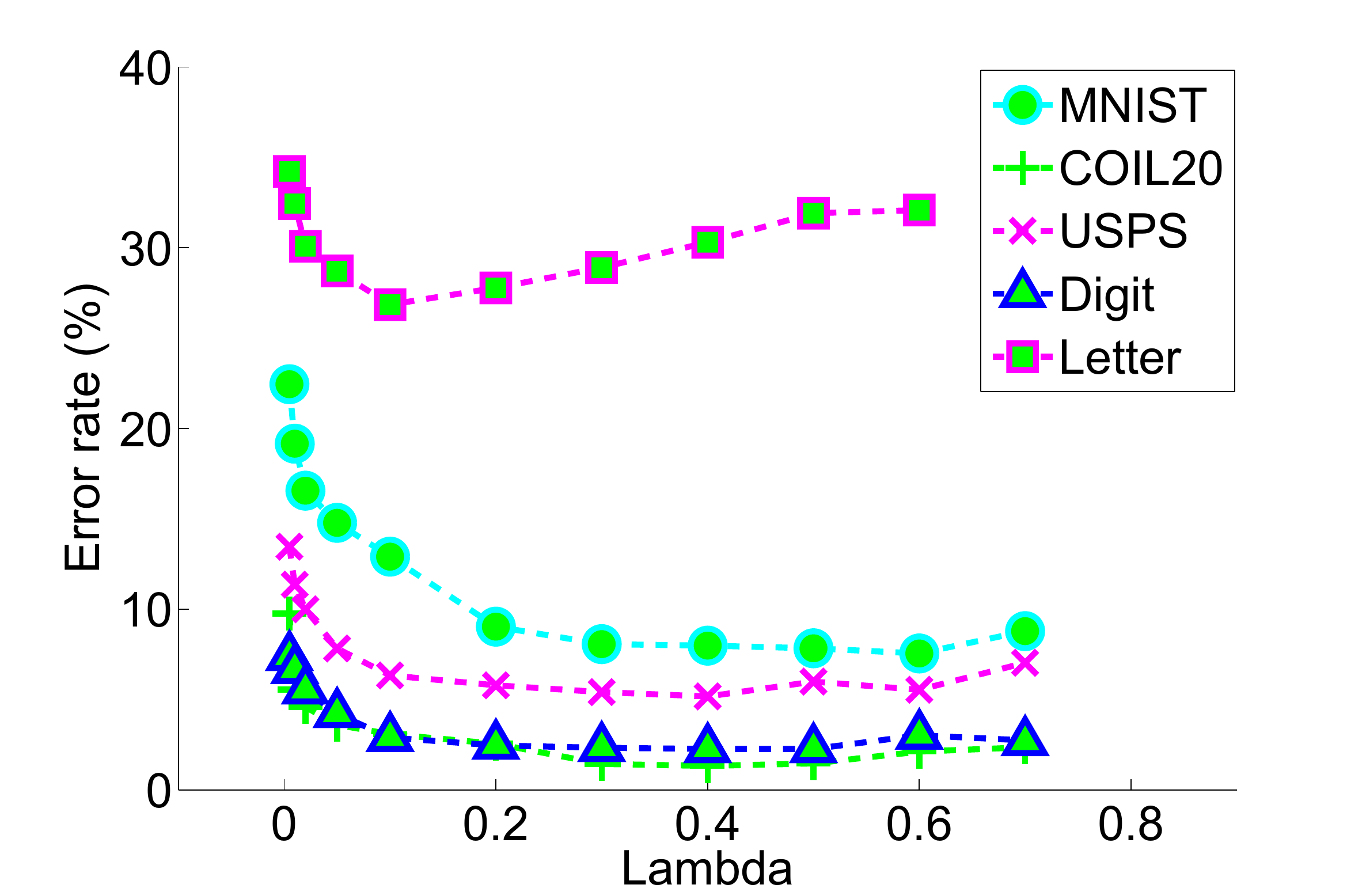}
                \caption{LSGC + SP (error rate (\%) vs. lambda)}
                \label{fig:lambdaEffect}
        \end{subfigure}       
         \caption{Classification performance of LSGC with different value of $k$-nn and $lambda$.}
         \label{fig:paramEffects}
\end{figure}

In Figure \ref{fig:paramEffects}, we show the influence of different choices of $\lambda$ and $k$ on the classification performance of the LSGC. It is shown that when the parameter ensures that the coding vectors of the base method are local enough, the algorithm performance is not too sensitive to the its exact value. This prior knowledge help us to decrease the number of test values in cross validation and increase the speed of the algorithm.

\begin{table*}[!th]
\caption{object recognition error rates (\%) for different similarity, locality and sparsity based coding methods on COIL20. The number in the parentheses are the $t$-step size retained by the cross validation.}
\begin{center}
\begin{tabular}{|c||c||c|c||c|c|c||c|c|c|}
\hline
\multirow{2}{*}{\#Training} & \multirow{2}{*}{OR} & \multicolumn{2}{|c||}{Similarity Based} &  \multicolumn{3}{|c||}{Locality Based} &  \multicolumn{3}{|c|}{Sparsity Based}\\ 
\cline{3-10} & & SAC & LSGC+SAC & LCC & LLC  & LSGC+LLC & SC & RSC & LSGC+SC\\
\hline
\hline
5 & $17.07$ & $16.87$ & $\mathbf{2.66}(11)$ & $19.50$ & $18.03$ & $\underline{4.41}(8)$ & $18.53$ & $13.68$ & $\underline{6.46}(15)$ \\
\hline
10 & $9.96$ & $9.11$ & $\mathbf{1.20}(11)$ & $12.10$ & $9.34$ & $\underline{2.32}(8)$ & $9.15$ &  $7.35$ & $\underline{3.00}(15)$ \\
\hline
20 & $4.83$ & $3.98$ & $\mathbf{0.53}(8)$ & $4.67$ & $3.15$ & $\underline{1.50}(5)$ & $3.16$ & $2.55$ & $\underline{1.45}(11)$ \\
\hline
30 & $3.03$ & $2.16$ & $\mathbf{0.51}(5)$ & $2.70$ & $0.89$ & $\underline{0.63}(2)$ & $1.36$ & $1.07$ & $\underline{0.85}(3)$ \\
\hline
\end{tabular}
\end{center}
\label{c_COIL20}
\end{table*}

\begin{table*}[!th]
\caption{digit recognition error rates (\%) for different similarity, locality and sparsity based coding methods on Digit. The number in the parentheses are the $t$-step size retained by the cross validation.}
\begin{center}
\begin{tabular}{|c||c||c|c||c|c|c||c|c|c|}
\hline
\multirow{2}{*}{\#Training} & \multirow{2}{*}{OR} & \multicolumn{2}{|c||}{Similarity Based} &  \multicolumn{3}{|c||}{Locality Based} &  \multicolumn{3}{|c|}{Sparsity Based}\\ 
\cline{3-10} & & SAC & LSGC+SAC & LCC & LLC  & LSGC+LLC & SC & RSC & LSGC + SC\\
\hline
\hline

5 & $14.78$ & $8.39$ & $\mathbf{3.35}(8)$ & $22.19$ & $19.36$ & $\underline{3.54}(11)$ & $16.62$ & $12.10$ & $\underline{3.40}(15)$ \\
\hline
10 & $10.69$ & $5.47$ & $\underline{3.10}(5)$ & $15.58$ & $13.93$ & $\underline{2.90}(8)$ & $9.45$ & $7.90$ & $\mathbf{2.89}(8)$ \\
\hline
20 & $7.79$ & $3.77$ & $\mathbf{2.03}(8)$ & $9.68$ & $6.36$ & $\underline{2.42}(5)$ & $6.70$ & $4.48$ & $\underline{2.08}(15)$ \\
\hline
30 & $6.82$ & $3.02$ & $\underline{2.55}(3)$ & $6.90$ & $5.03$ & $\underline{2.44}(5)$ & $4.15$ & $3.74$ & $\mathbf{2.02}(8)$ \\
\hline
60 & $5.38$ & $2.25$ & $\underline{2.00}(8)$ & $4.59$ & $3.22$ & $\underline{1.68}(5)$ & $3.16$ & $2.91$ & $\mathbf{1.61}(5)$ \\
\hline
100 & $4.64$ & $1.87$ & $1.85(1)$ & $2.97$ & $2.53$ & $\underline{1.82}(2)$ & $2.51$ & $2.04$ & $\mathbf{1.36}(2)$ \\
\hline
\end{tabular}
\label{c_Digit}
\end{center}
\end{table*}

\begin{table*}[!th]
\caption{letter recognition error rates (\%) for different similarity, locality and sparsity based coding methods on Letter. The number in the parentheses are the $t$-step size retained by the cross validation.}
\begin{center}
\begin{tabular}{|c||c||c|c||c|c|c||c|c|c|}
\hline
\multirow{2}{*}{\#Training} & \multirow{2}{*}{OR} & \multicolumn{2}{|c||}{Similarity Based} &  \multicolumn{3}{|c||}{Locality Based} &  \multicolumn{3}{|c|}{Sparsity Based}\\ 
\cline{3-10} & & SAC & LSGC+SAC & LCC & LLC  & LSGC+LLC & SC & RSC & LSGC + SC\\
\hline
\hline
5 & $49.78$ & $51.95$ & $\underline{50.66}(5)$ & $57.87$ & $49.16$ & $\mathbf{48.21}(2)$ & $54.79$ & $57.57$ & $\underline{48.38}(5)$ \\
\hline
10 & $42.57$ & $39.18$ & $39.81(1)$ & $47.55$ & $38.05$ & $38.36(1)$ & $43.38$ & $\mathbf{37.47}$ & $37.69(3)$ \\
\hline
20 & $37.69$ & $28.31$ & $28.04(1)$ & $36.60$ & $27.93$ & $27.99(1)$ & $32.87$ & $\mathbf{27.63}$ & $27.83(2)$ \\
\hline
30 & $35.94$ & $22.64$ & $\mathbf{22.31}(1)$ & $31.22$ & $24.88$ & $24.82(1)$ & $26.63$ & $23.25$ & $\underline{22.52}(2)$ \\
\hline
60 & $33.54$ & $\mathbf{16.00}$ & $16.03(1)$ & $23.83$ & $18.00$ & $18.06(1)$ & $20.18$ & $\underline{17.28}$ & $17.43(2)$ \\
\hline
100 & $32.74$ & $12.59$ & $\mathbf{12.44}(1)$ & $19.54$ & $14.77$ & $14.68(1)$ & $17.06$ & $14.76$ & $\underline{14.47}(2)$ \\
\hline
\end{tabular}
\end{center}
\label{c_Letter}
\end{table*}

\begin{table*}[!th]
\caption{digit recognition error rates (\%) for different similarity, locality and sparsity based coding methods on MNIST. The number in the parentheses are the $t$-step size retained by the cross validation.}
\begin{center}
\begin{tabular}{|c||c||c|c||c|c|c||c|c|c|}
\hline

\multirow{2}{*}{\#Training} & \multirow{2}{*}{OR} & \multicolumn{2}{|c||}{Similarity Based} &  \multicolumn{3}{|c||}{Locality Based} &  \multicolumn{3}{|c|}{Sparsity Based}\\ 
\cline{3-10} & & SAC & LSGC+SAC & LCC & LLC  & LSGC+LLC & SC & RSC & LSGC + SC\\
\hline
\hline
5 & $33.03$ & $23.17$ & $\mathbf{10.95}(3)$ & $34.86$ & $40.38$ & $\underline{11.31}(5)$ & $42.53$ & $27.31$ & $\underline{13.78}(8)$ \\
\hline
10 & $25.96$ & $15.36$ & $\mathbf{9.20}(15)$ & $24.47$ & $29.10$ & $\underline{9.97}(8)$ & $27.49$ & $24.60$ & $\underline{9.36}(11)$ \\
\hline
20 & $20.97$ & $10.91$ & $\underline{7.49}(2)$ & $17.59$ & $19.77$ & $\mathbf{7.40}(3)$ & $15.93$ & $11.72$ & $\underline{7.55}(8)$ \\
\hline
30 & $18.97$ & $8.88$ & $\mathbf{6.51}(3)$ & $14.17$ & $15.62$ & $\underline{6.82}(8)$ & $13.73$ & $9.99$ & $\underline{7.19}(8)$ \\
\hline
60 & $16.09$ & $7.01$ & $\mathbf{6.12}(3)$ & $10.50$ & $11.24$ & $\underline{6.17}(3)$ & $9.08$ & $6.95$ & $\underline{6.34}(3)$ \\
\hline
100 & $14.68$ & $6.16$ & $\underline{6.00}(2)$ & $8.48$ & $9.40$ & $\underline{5.96}(5)$ & $7.11$ & $5.78$ & $\mathbf{5.61}(3)$ \\
\hline
\end{tabular}
\end{center}
\label{c_MNIST}
\end{table*}

\begin{table*}[!th]
\caption{digit recognition error rates (\%) for different similarity, locality and sparsity based coding methods on USPS. The number in the parentheses are the $t$-step size retained by the cross validation.}
\begin{center}
\begin{tabular}{|c||c||c|c||c|c|c||c|c|c|}
\hline
\multirow{2}{*}{\#Training} & \multirow{2}{*}{OR} & \multicolumn{2}{|c||}{Similarity Based} &  \multicolumn{3}{|c||}{Locality Based} &  \multicolumn{3}{|c|}{Sparsity Based}\\ 
\cline{3-10} & & SAC & LSGC+SAC & LCC & LLC  & LSGC+LLC & SC & RSC & LSGC + SC\\
\hline
\hline
5 & $22.04$ & $17.91$ & $\underline{12.50}(3)$ & $30.35$ & $27.13$ & $\underline{9.23}(5)$ & $25.67$ & $17.67$ & $\mathbf{5.94}(15)$ \\
\hline
10 & $15.73$ & $12.00$ & $\underline{7.64}(3)$ & $21.31$ & $18.95$ & $\underline{7.56}(5)$ & $14.94$ & $12.60$ & $\mathbf{5.46}(15)$ \\
\hline
20 & $11.97$ & $9.21$ & $\underline{7.12}(2)$ & $13.49$ & $14.41$ & $\underline{7.11}(2)$ & $10.25$ & $8.71$ & $\mathbf{5.42}(11)$ \\
\hline
30 & $10.42$ & $7.28$ & $\underline{6.21}(3)$ & $10.51$ & $9.92$ & $\mathbf{5.41}(3)$ & $8.40$ & $7.10$ & $\underline{6.11}(3)$ \\
\hline
60 & $8.62$ & $5.95$ & $6.03(1)$ &  $7.60$ & $7.17$ & $\underline{5.26}(2)$ & $6.49$ & $5.41$ & $\mathbf{4.71}(15)$ \\
\hline
100 & $7.66$ & $5.10$ & $5.29(1)$ & $5.90$ & $5.82$ & $\underline{4.79}(2)$ & $5.62$ & $4.93$ & $\mathbf{3.87}(2)$ \\
\hline
\end{tabular}
\end{center}
\label{c_USPS}
\end{table*}

\subsection{Image Clustering}
As another application of coding algorithms, here we investigate the effectiveness of different methods on image clustering. The comparisons are made on COIL20 and Digit datasets which are well-known for this application. Different coding methods are studied: LLC, Sparse Coding (SC), Regularized Sparse Coding (RSC) and LSGC. In the experiments, first PCA is applied to reduce the dimensionality while preserving 98 percent of the data energy, then k-means is performed in each coding space to cluster the data. Also k-means on the raw data as a baseline, and N-CUT \cite{shi00} as a famous spectral clustering algorithm are reported in the following experiments.

\subsubsection{Evaluation Methods} 
To evaluate the clustering performance, Normalized Mutual Information (NMI) and Accuracy (AC), as two standard criteria, are used. In both criteria, the true label of points are considered as the true cluster labels and the computed cluster labels are compared with them. Assume that $C = \{ c_1,\hdots,c_M \}$ is the set of true clusters and $C' = \{c'_1,\hdots,c'_M\}$ is the set of computed clusters, labeled with their indices. e.g. $c_i$ contains all the points in the $i$'th class and $c'_i$ is the set of points in the $i$'th computed cluster. The mutual information between these two sets is computed as follows:
\begin{equation}
MI(C,C') = \sum_{c_i \in C , c'_j \in C'}p(c_i,c'_j) \log_2\frac{p(c_i,c'_j)}{p(c_i)\times p(c'_j)},
\end{equation}
in which $p(c_i)$ and $p(c'_j)$ are the probabilities that a selected point be in the clusters $c_i$ and $c'_j$ respectively and $p(c_i,c'_j) $ is the probability that a selected point belongs clusters $c_i$ and $c'_j$ simultaneously. To map the computed mutual information into $[0,1]$ interval a normalization can be applied:
\begin{equation}
NMI(C,C') = \frac{MI(C,C')}{\max({H(C),H(C')})}
\end{equation}
here $H(.)$ is the standard entropy of the corresponding set. It can be easily verified that if $C$ and $C'$ match completely, regardless of their labels, $NMI(C,C')$ will be equal to one.

For the AC criterion, first each computed cluster is labeled with one of the classes in the dataset with Kuhn-Munkres algorithm \cite{lovasz}. If we denote $l(\mathbf{x})$ and $l'(\mathbf{x})$ as the functions that return the true and computed cluster label of a given point $\mathbf{x}$ respectively and $map(l'(\mathbf{x}))$ is the Kuhn-Munkres mapping function, which maps the value of $l'(\mathbf{x})$ into a value in the range of $l$ function, then the accuracy of a clustering algorithm can be defined as follows:
\begin{equation}
AC = \sum_{\mathbf{x} \in \mathbf{X}} \frac{\mathcal{I}\big(l(\mathbf{x})==map(l'(\mathbf{x})\big)}{N},
\end{equation}   
in which $\mathbf{X}$ is the set of all images in the dataset, $N$ is the total number of images in the dataset and $\mathcal{I}(.)$ is the indicator function which returns one if its condition is met.

\subsubsection{Experimental Results}
In the experiments, parameters for each method are selected empirically to have the best performance. However, for our methods we fixed the $k$-nn parameter for the core LLC to 3 and the $\lambda$ parameter for our core sparse encoder to $0.3$ for both datasets. Each experiment is performed $50$ times and the average performance is reported. Also, in each run the k-means algorithm is performed $20$ times with different random initialization and the best result is reported for each method. Tables \ref{tab:clustering_COIL} and \ref{tab:clustering_Digit} show the NMI and AC measure of the methods for different number of clusters on COIL20 and Digit datasets respectively. For each row in the tables, the mentioned number of classes are selected randomly from the dataset in each run. As can be seen in these tables, our modified LLC and SC methods outperform others by a vast margin. 

\begin{table*}[] 

 \caption{Clustering results for different coding methods on the COIL20 dataset.}
  \label{tab:clustering_COIL}
 \begin{center} 
 
 \begin{tabular}{|c||c||c|c||c|c||c|c|c|} 
 \hline 
 \multirow{2}{*}{\#Clusters} & \multirow{2}{*}{Criterion} & \multicolumn{2}{|c||}{Base Methods} &  \multicolumn{2}{|c||}{Locality Based} &  \multicolumn{3}{|c|}{Sparsity Based}\\ 
 \cline{3-9}
 & & Kmeans& NCUT & LLC & LSGC+LLC & SP  & RSP & LSGC+SP \\
 
\hline \hline 
 
\multirow{2}{*}{2} & NMI & \underline{80.79} & 77.85 & 81.04  & \bf{98.48}  & 74.05  & 76.77  &  \underline{98.42}   \\ 
\cline{2-9}
 & AC & \underline{93.76} & 92.46 & 93.46 & \bf{99.40} & 89.83  & 91.92 &  \underline{99.36}  \\ 
 \hline 
 
\multirow{2}{*}{4} & NMI & 76.60  & \underline{88.36} & 82.24 & \bf{98.09} & 77.54  & 78.70 &  \underline{93.73} \\ 
\cline{2-9}
& AC & 82.50 & \underline{90.65} & 86.99 & \bf{98.49} & 81.24 & 83.98 &  \underline{93.78}  \\ 
 
\hline

\multirow{2}{*}{6} & NMI & 76.29 & \underline{85.88}& 78.93&  \bf{97.75} & 76.97  & 79.60  & \underline{90.28}  \\ 
 
 \cline{2-9} 
 & AC & 76.32 & \underline{83.53} & 78.58 & \bf{97.21} & 74.40 & 78.80 &  \underline{84.72}   \\ 
  
 \hline 
 
\multirow{2}{*}{8} & NMI & 76.23 & \underline{87.46} & 74.40  & \bf{95.32} & 73.06 & 79.03  &  \underline{88.18}  \\ 
 
 \cline{2-9}
 & AC & 73.28 & \underline{83.57} & 71.20 &  \bf{92.28}  & 66.97 & 76.12 & \underline{79.72}    \\ 
  
  \hline
 
\multirow{2}{*}{10} & NMI & 78.08 & \underline{83.19} & 76.07 & \bf{94.74} & 73.11  & 78.57 &  \underline{88.24}  \\ 
 
 \cline{2-9} 
 & AC & 73.74 & \underline{77.13} & 69.36 & \bf{89.21} &  65.25  & 73.79  &  \underline{77.67} \\ 
  
 \hline
 
\multirow{2}{*}{12} & NMI & 79.45 & \underline{80.33} & 75.67 & \bf{93.35} & 73.75 & 76.35 &  \underline{87.24}  \\ 
 
\cline{2-9} 
 & AC & 72.64  & \underline{73.07} & 68.08 & \bf{86.14} & 64.63 &  69.15 & \underline{76.01}\\ 
  
  \hline 
 
\multirow{2}{*}{14} & NMI &  76.20 & \underline{81.47}  & 75.55  & \bf{93.21} & 74.96 & 77.56 &  \underline{87.60}   \\ 
 
 \cline{2-9}
 & AC & 68.59& \underline{73.03}& 65.83 & \bf{85.60} & 64.95 & 68.71&  \underline{74.54} \\ 
  
  \hline
 
\multirow{2}{*}{16} & NMI & 77.41 & \underline{77.81}  & 75.72 & \bf{93.21}  & 73.46  & 76.62 &  \underline{85.79}   \\ 
 
 \cline{2-9} 
 & AC & \underline{68.84} & 65.99 & 64.97 & \bf{84.77}  & 62.37  & 66.56 &  \underline{71.38}  \\ 
  
  \hline
 
\multirow{2}{*}{18} & NMI & 76.69  & \underline{77.76} & 74.76  & \bf{91.38}  & 74.18  & 76.91  &  \underline{85.26} \\ 
 
 \cline{2-9}
 & AC & \underline{65.53} & 65.14 & 63.18 & \bf{80.97} & 62.46  & 66.00 &  \underline{69.48}  \\ 
  
  \hline
 
\multirow{2}{*}{20} & NMI & 77.10 & \underline{79.87} & 74.63  & \bf{91.86}  & 73.42 & 76.55  &  \underline{84.95}  \\ 
 
\cline{2-9} 
 & AC & 65.35 & \underline{65.91} & 62.46 & \bf{81.85}  & 60.65 & 64.57 &  \underline{68.14}  \\ 
  
  \hline
\hline 
\multirow{2}{*}{Avg} & NMI & 77.48 & \underline{82.00}  & 76.90 &  \bf{94.74}  & 74.45  & 77.67 & \underline{88.97}  \\ 
 
 \cline{2-9} 
 & AC & 74.06  & \underline{77.05} & 72.41 & \bf{89.59} & 69.28  & 73.96 &  \underline{79.48}  \\ 
  
  \hline
 
\end{tabular} 
 
 \end{center} 
 
 \end{table*}

  
 \begin{table*}[] 
 
  \caption{Clustering results for different coding methods on the Digit dataset.}
   \label{tab:clustering_Digit}
  \begin{center} 
  
  \begin{tabular}{|c||c||c|c||c|c||c|c|c|} 
   \hline 
   \multirow{2}{*}{\#Clusters} & \multirow{2}{*}{Criterion} & \multicolumn{2}{|c||}{Base Methods} &  \multicolumn{2}{|c||}{Locality Based} &  \multicolumn{3}{|c|}{Sparsity Based}\\ 
   \cline{3-9}
   & & Kmeans& NCUT & LLC & LSGC+LLC & SP  & RSP & LSGC+SP \\
   
  \hline \hline 
  
 \multirow{2}{*}{2} & NMI& \underline{83.15}& 81.55 & 83.84  & \bf{93.03}  & 83.66   & 86.55& \underline{86.74} \\ 
  \cline{2-9}
  & AC& 95.76 & \underline{95.91} & 96.19  & \bf{98.43} & 94.37  & \underline{97.19}& 96.64  \\ 
  \hline 
  
 \multirow{2}{*}{3} & NMI& 79.88& \underline{85.58} & 83.79   & \bf{95.17} & 86.52   & 84.84 & \underline{91.52} \\ 
  \cline{2-9}
  & AC & 93.21& \underline{96.07}  & 94.45    & \bf{99.04} & 94.50  & 94.88 & \underline{97.17} \\
  \hline 
  
 \multirow{2}{*}{4} & NMI & 79.10& \underline{85.06}  & 82.69  & \bf{93.32}& 81.20 & 85.09 & \underline{88.89}  \\ 
  \cline{2-9}
  & AC& 90.27 & \underline{94.53} & 91.73 & \bf{97.62} & 88.46  & 93.07& \underline{94.58} \\ 
  \hline 
  
 \multirow{2}{*}{5} & NMI & 79.30& \underline{82.71} & 80.69 & \bf{90.83} & 83.08 & 81.14 & \underline{86.54} \\ 
  \cline{2-9}
  & AC& 88.56 & \underline{92.75}  & 88.84   & \bf{95.45}& 88.99 & 89.30 & \underline{92.07}  \\ 
  \hline 
  
 \multirow{2}{*}{6} & NMI & 75.05& \underline{81.40}  & 77.67  & \bf{88.60} & 80.33  & 78.45  & \underline{85.87} \\ 
  \cline{2-9}
  & AC& 83.85& \underline{90.36}  & 85.97 & \bf{94.00} & 86.37   & 84.80& \underline{90.20} \\ 
  \hline 
  
 \multirow{2}{*}{7} & NMI & 75.64 & \underline{77.81} & 78.04    & \bf{94.58}& 80.83  & 79.33 & \underline{86.81} \\ 
  \cline{2-9}
  & AC& 82.73 & \underline{85.79} & 85.21   & \bf{98.01} & 85.80 & 85.00 & \underline{90.06}\\
  \hline 
  
 \multirow{2}{*}{8} & NMI & 73.65& \underline{76.94} & 77.26   & \bf{94.90}& 78.23& 77.70 & \underline{85.09} \\ 
  \cline{2-9}
  & AC & 79.43& \underline{84.24} & 83.77    & \bf{97.97} & 81.77  & 82.58& \underline{87.17} \\ 
  \hline 
  
 \multirow{2}{*}{9} & NMI & 74.32 & \underline{77.14}  & 74.03  & \bf{85.74} & 77.75   & 78.26& \underline{84.62}\\ 
  \cline{2-9}
  & AC & 79.33 & \underline{84.65}  & 78.29   & \bf{86.02} & 80.28 & 82.68 & \underline{85.37} \\ 
  \hline 
  
 \multirow{2}{*}{10} & NMI & 74.83& \underline{78.89} & 71.85  & \bf{89.33}  & 78.28  & 77.87& \underline{83.79} \\ 
  \cline{2-9}
  & AC  & 79.21 & \underline{87.98} & 74.48    & \bf{91.62}& 81.12  & 83.02& \underline{83.83}\\
  \hline 
  \hline
 \multirow{2}{*}{Avg} & NMI & 77.21 & \underline{80.79}& 78.87   & \bf{91.72} & 81.10 & 81.02 & \underline{86.65}\\ 
  \cline{2-9}
  & AC & 85.82 & \underline{90.25} & 86.55   & \bf{95.35}  & 86.85  & 88.06 & \underline{90.79} \\
  \hline 
  
 \end{tabular} 
  
  \end{center} 
  
  \end{table*}

\section{Conclusion}
\label{lab:conclusions}
In this paper we present a method called LSGC that considers the coding of each basis as a way to capture the underlying structure of data and exploits it to make the coding coefficients more accurate. To put in another word, as illustrated in Figure \ref{fig:contributions}, compared to the conventional coding schemes which are based on the Euclidean distance, our coding coefficients are assigned according to the similarity measure that changes smoothly over the data manifold. In a theoretical point of view, the linear kernel in our coding space is a approximation to the diffusion kernel which is a well-known kernel in manifold learning literature. The experimental results on different learning tasks show the effectiveness of the method.

The assumption that we implicitly have in our method is that the bases imitate the structure of the data points. Thus, the value of $\tilde{p}^{2t}(\mathbf{x,y})$ on the graph containing the bases as intermediate nodes approximates this transition probability on the graph over all the data points. Therefore, it remains an open issue to theoretically bound the approximation error of $\tilde{p}^{2t}(\mathbf{x,y})$ on the graph contains bases which its nodes are learned using different dictionary methods compared to the graph over the original data points.


%

\appendix[Proof of Thoerem 1]
\begin{theorem}
$\frac{\mathbf{r}^{2t} (\mathbf{x}, \mathbf{y})}{\tilde{p}^{2t} (\mathbf{x}, \mathbf{y})}$ converges to zero at the rate of $\mathcal{O}(K^{-1})$ where $K$ is the number of visual bases in the graph.
\begin{IEEEproof}
For the paths that meet $\mathbf{x}$ more than once, there is a step $1 \le j < 2t$ at which the random walker returns to $\mathbf{x}$ (Similar proof can be driven by substituting $\mathbf{y}$ instead of $\mathbf{x}$ if $\mathbf{y}$ is met more than once), so: 
\begin{equation}
\label{conv_equation}
\frac{\mathbf{r}^{2t} (\mathbf{x}, \mathbf{y})}{\tilde{p}^{2t} (\mathbf{x}, \mathbf{y})} \le \sum_{j = 1}^{2t}  \frac{\tilde{p}^{j}(\mathbf{x}, \mathbf{x}) \tilde{p}^{2t-j}(\mathbf{x}, \mathbf{y})}{\tilde{p}^{2t}(\mathbf{x}, \mathbf{y})}
\end{equation}
This inequality holds since the paths which return to $\mathbf{x}$ more than once are enumerated several times in the above sum. By unrolling $\tilde{p}^{2t}(\mathbf{x}, \mathbf{y})$  using the Chapman-Kolmogorov equation, we have:

\begin{equation}
\label{low_equ}
\begin{split}
\frac{\mathbf{r}^{2t} (\mathbf{x}, \mathbf{y})}{\tilde{p}^{2t} (\mathbf{x}, \mathbf{y})} &\le \sum_{j = 1}^{2t}  \frac{\tilde{p}^{j}(\mathbf{x}, \mathbf{x}) \tilde{p}^{2t-j}(\mathbf{x}, \mathbf{y})}{\sum_{i=1}^{K+2} \tilde{p}^{j}(\mathbf{x}, \mathbf{z_i})\tilde{p}^{2t-j}(\mathbf{z_i}, \mathbf{y})} \\
&= \sum_{j = 1}^{2t}  \Big( \sum_{i=1}^{K+2} \frac{\tilde{p}^{j}(\mathbf{x}, \mathbf{z_i})}{\tilde{p}^{i}(\mathbf{x}, \mathbf{x})} \frac{\tilde{p}^{2t-j}(\mathbf{z_i}, \mathbf{y})}{\tilde{p}^{2t-i}(\mathbf{x}, \mathbf{y})} \Big)^{-1} 
\end{split}
\end{equation}
where the inner sum is over $\mathbf{x}$, $\mathbf{y}$, and $K$ other bases. The idea is to find a lower bound for ${\tilde{p}^{j}(\mathbf{m}, \mathbf{z})}/{\tilde{p}^{j}(\mathbf{n}, \mathbf{z})}$  where $\mathbf{m}$, $\mathbf{n}$, and $\mathbf{z}$ are arbitrary points in the graph, and substitute it in (\ref{low_equ}). To do so, we start with the following trivial inequality:
\begin{equation}
\norm{\mathbf{m} - \mathbf{z}}^2 - \mathfrak{L}^2 \le \norm{\mathbf{n}-\mathbf{z}}^2 \le \norm{\mathbf{m}-\mathbf{z}}^2+\mathfrak{L}^2
\end{equation}
where $\mathfrak{L}$ is the diameter of the sphere containing all the sampled data points. The limited support of the data distribution guarantees the existence of $\mathfrak{L}$. So we have:
\begin{equation}
\label{exp_equ}
\begin{split}
& \exp{(-\frac{\norm{\mathbf{m}-\mathbf{z}}^2}{2 \sigma^2})} \exp{(-\frac{\mathfrak{L}^2}{2 \sigma^2})} \le \exp{(-\frac{\norm{\mathbf{n}-\mathbf{z}}^2}{2 \sigma^2})} \\
\le &\exp{(-\frac{\norm{\mathbf{m}-\mathbf{z}}^2}{2 \sigma^2})} \exp{(\frac{\mathfrak{L}^2}{2 \sigma^2})}.
\end{split}
\end{equation}
For a one step random walk on the graph, in which the edges are weighted by a Gaussian kernel we have:
\begin{equation}
\label{p1_equ}
\begin{split}
\tilde{p}^1(\mathbf{n}, \mathbf{z}) &= \frac{\mathbf{R}(\mathbf{n}, \mathbf{z})}{\sqrt{d(\mathbf{n})} \sqrt{d(\mathbf{z})}} \\
&= \frac{\exp{(-\frac{\norm{\mathbf{n}-\mathbf{z}}^2}{2 \sigma^2})}}{\sqrt{\sum_{k=1}^{K+2} \exp{(-\frac{\norm{\mathbf{n}-\mathbf{y_k}}^2}{2 \sigma^2})} } \sqrt{d(\mathbf{z})}}
\end{split}
\end{equation}
To obtain a lower bound we use the bounds in (\ref{exp_equ}) and substitute the lower bound in its numerator and the upper bound in its denominator:
\begin{equation}
\begin{split}
\tilde{p}^1(\mathbf{n}, \mathbf{z}) &\ge \frac{\exp{(-\frac{\norm{\mathbf{m}-\mathbf{z}}^2}{2 \sigma^2})} \exp{(-\frac{\mathfrak{L}^2}{2 \sigma^2})} }{\sqrt{\sum_{k=1}^{K+2} \exp{(-\frac{\norm{\mathbf{m}-\mathbf{y_k}}^2}{2 \sigma^2}) \exp{(\frac{\mathfrak{L}^2}{2 \sigma^2})}} } \sqrt{d(\mathbf{z})}} \\
&= \frac{\exp{(-\frac{\norm{\mathbf{m}-\mathbf{z}}^2}{2 \sigma^2})}}{\sqrt{\sum_{k=1}^{K+2} \exp{(-\frac{\norm{\mathbf{m}-\mathbf{y_k}}^2}{2 \sigma^2}) } } \sqrt{d(\mathbf{z})}}  \times \frac{\exp{(-\frac{\mathfrak{L}^2}{2 \sigma^2})}}{ \exp{(\frac{\mathfrak{L}^2}{4 \sigma^2})}} \\
&= \tilde{p}^1(\mathbf{m}, \mathbf{z}) \exp{(- \frac{3 \mathfrak{L}^2}{4 \sigma^2})}
\end{split}
\end{equation}
Now we use this to bound $j$-step random walks:
\begin{equation}
\label{j_step_equ}
\begin{split}
\frac{\tilde{p}^{j}(\mathbf{n}, \mathbf{y})}{\tilde{p}^{j}(\mathbf{m}, \mathbf{y})} &= \frac{\sum_{l = 1}^{K+2} \tilde{p}^1(\mathbf{n}, \mathbf{y_l}) \tilde{p}^{j-1}(\mathbf{y_l}, \mathbf{y})}{\tilde{p}^j(\mathbf{m},\mathbf{y})} \\
&\ge  \frac{\exp{(- \frac{3 \mathfrak{L}^2}{4 \sigma^2})} \sum_{l = 1}^{K+2} \tilde{p}^1(\mathbf{m}, \mathbf{y_l}) \tilde{p}^{j-1}(\mathbf{y_l}, \mathbf{y})}{\tilde{p}^j(\mathbf{m},\mathbf{y})} \\
&= \exp{(- \frac{3 \mathfrak{L}^2}{4 \sigma^2})}
\end{split}
\end{equation}
By substituting (\ref{j_step_equ}) in (\ref{low_equ}) we have:
\begin{equation}
\begin{split}
\frac{\mathbf{r}^{2t} (\mathbf{x}, \mathbf{y})}{\tilde{p}^{2t} (\mathbf{x}, \mathbf{y})} &\le \sum_{j = 1}^{2t}  \Big( \sum_{i=1}^{K+2} \frac{\tilde{p}^{j}(\mathbf{x}, \mathbf{z_i})}{\tilde{p}^{i}(\mathbf{x}, \mathbf{x})} \frac{\tilde{p}^{2t-j}(\mathbf{z_i}, \mathbf{y})}{\tilde{p}^{2t-i}(\mathbf{x}, \mathbf{y})} \Big)^{-1} \\
&\le \sum_{j = 1}^{2t}  \Big( \sum_{i=1}^{K+2} 
\exp{(- \frac{3 \mathfrak{L}^2}{4 \sigma^2})} 
\exp{(- \frac{3 \mathfrak{L}^2}{4 \sigma^2})} \Big)^{-1} \\
&= \frac{t \exp{(\frac{3 \mathfrak{L}^2}{2 \sigma^2})}}{K+2}
\end{split}
\end{equation}
Therefore,
\begin{equation}
0 \le \frac{\mathbf{r}^{2t} (\mathbf{x}, \mathbf{y})}{\tilde{p}^{2t} (\mathbf{x}, \mathbf{y})} \le \frac{t \exp{(\frac{3 \mathfrak{L}^2}{2 \sigma^2})}}{K+2}
\end{equation}

\end{IEEEproof}
\end{theorem}




\ifCLASSOPTIONcaptionsoff
  \newpage
\fi



\bibliographystyle{IEEEtran}
\bibliography{IEEEabrv.bib,egbibNEW.bib}

\begin{thebibliography}{10}
\providecommand{\url}[1]{#1}
\csname url@samestyle\endcsname
\providecommand{\newblock}{\relax}
\providecommand{\bibinfo}[2]{#2}
\providecommand{\BIBentrySTDinterwordspacing}{\spaceskip=0pt\relax}
\providecommand{\BIBentryALTinterwordstretchfactor}{4}
\providecommand{\BIBentryALTinterwordspacing}{\spaceskip=\fontdimen2\font plus
\BIBentryALTinterwordstretchfactor\fontdimen3\font minus
  \fontdimen4\font\relax}
\providecommand{\BIBforeignlanguage}[2]{{%
\expandafter\ifx\csname l@#1\endcsname\relax
\typeout{** WARNING: IEEEtran.bst: No hyphenation pattern has been}%
\typeout{** loaded for the language `#1'. Using the pattern for}%
\typeout{** the default language instead.}%
\else
\language=\csname l@#1\endcsname
\fi
#2}}
\providecommand{\BIBdecl}{\relax}
\BIBdecl

\bibitem{lcc09}
K.~Yu, T.~Zhang, and Y.~Gong, ``Nonlinear learning using local coordinate
  coding,'' \emph{Advances in Neural Information Processing Systems}, vol.~22,
  pp. 2223--2231, 2009.

\bibitem{cevher2008compressive}
V.~Cevher, A.~Sankaranarayanan, M.~F. Duarte, D.~Reddy, R.~G. Baraniuk, and
  R.~Chellappa, ``Compressive sensing for background subtraction,'' in
  \emph{Computer Vision--ECCV 2008}.\hskip 1em plus 0.5em minus 0.4em\relax
  Springer, 2008, pp. 155--168.

\bibitem{dikmen2008robust}
M.~Dikmen and T.~S. Huang, ``Robust estimation of foreground in surveillance
  videos by sparse error estimation,'' in \emph{Pattern Recognition, 2008. ICPR
  2008. 19th International Conference on}.\hskip 1em plus 0.5em minus
  0.4em\relax IEEE, 2008, pp. 1--4.

\bibitem{yang2008image}
J.~Yang, J.~Wright, T.~Huang, and Y.~Ma, ``Image super-resolution as sparse
  representation of raw image patches,'' in \emph{Computer Vision and Pattern
  Recognition, 2008. CVPR 2008. IEEE Conference on}.\hskip 1em plus 0.5em minus
  0.4em\relax IEEE, 2008, pp. 1--8.

\bibitem{liu2011robust}
B.~Liu, J.~Huang, L.~Yang, and C.~Kulikowsk, ``Robust tracking using local
  sparse appearance model and k-selection,'' in \emph{Computer Vision and
  Pattern Recognition (CVPR), 2011 IEEE Conference on}.\hskip 1em plus 0.5em
  minus 0.4em\relax IEEE, 2011, pp. 1313--1320.

\bibitem{wang2013online}
D.~Wang, H.-C. Lu, and M.-H. Yang, ``Online object tracking with sparse
  prototypes,'' 2013.

\bibitem{zhang2012sparse}
S.~Zhang, H.~Yao, X.~Sun, and X.~Lu, ``Sparse coding based visual tracking:
  review and experimental comparison,'' \emph{Pattern Recognition}, 2012.

\bibitem{ganesh2009robust}
A.~Ganesh, S.~S. Sastry, and Y.~Ma, ``Robust face recognition via sparse
  representation,'' \emph{IEEE TRANSACTIONS ON PATTERN ANALYSIS AND MACHINE
  INTELLIGENCE}, vol.~31, no.~2, p.~1, 2009.

\bibitem{bow04}
G.~Csurka, C.~Dance, L.~Fan, J.~Willamowski, and C.~Bray, ``Visual
  categorization with bags of keypoints,'' in \emph{Workshop on statistical
  learning in computer vision, ECCV}, vol.~1, 2004, p.~22.

\bibitem{llc10}
J.~Wang, J.~Yang, K.~Yu, F.~Lv, T.~Huang, and Y.~Gong, ``Locality-constrained
  linear coding for image classification,'' in \emph{Computer Vision and
  Pattern Recognition (CVPR), 2010 IEEE Conference on}.\hskip 1em plus 0.5em
  minus 0.4em\relax IEEE, 2010, pp. 3360--3367.

\bibitem{sac10}
J.~C. van Gemert, C.~J. Veenman, A.~W.~M. Smeulders, and J.~M. Geusebroek,
  ``Visual word ambiguity,'' \emph{IEEE Transactions on Pattern Analysis and
  Machine Intelligence}, vol.~32, no.~7, pp. 1271--1283, 2010.

\bibitem{scspm09}
J.~Yang, K.~Yu, Y.~Gong, and T.~Huang, ``Linear spatial pyramid matching using
  sparse coding for image classification,'' in \emph{Computer Vision and
  Pattern Recognition, 2009. CVPR 2009. IEEE Conference on}.\hskip 1em plus
  0.5em minus 0.4em\relax Ieee, 2009, pp. 1794--1801.

\bibitem{bowpos06}
A.~Agarwal and B.~Triggs, ``Recovering 3d human pose from monocular images,''
  \emph{Pattern Analysis and Machine Intelligence, IEEE Transactions on},
  vol.~28, no.~1, pp. 44--58, 2006.

\bibitem{wang2009multi}
C.~Wang, S.~Yan, L.~Zhang, and H.-J. Zhang, ``Multi-label sparse coding for
  automatic image annotation,'' in \emph{Computer Vision and Pattern
  Recognition, 2009. CVPR 2009. IEEE Conference on}.\hskip 1em plus 0.5em minus
  0.4em\relax IEEE, 2009, pp. 1643--1650.

\bibitem{olshausen1997sparse}
B.~A. Olshausen, D.~J. Field \emph{et~al.}, ``Sparse coding with an
  overcomplete basis set: A strategy employed by vi?'' \emph{Vision research},
  vol.~37, no.~23, pp. 3311--3326, 1997.

\bibitem{davis1997adaptive}
G.~Davis, S.~Mallat, and M.~Avellaneda, ``Adaptive greedy approximations,''
  \emph{Constructive approximation}, vol.~13, no.~1, pp. 57--98, 1997.

\bibitem{chen1989orthogonal}
S.~Chen, S.~A. Billings, and W.~Luo, ``Orthogonal least squares methods and
  their application to non-linear system identification,'' \emph{International
  Journal of control}, vol.~50, no.~5, pp. 1873--1896, 1989.

\bibitem{mallat1993matching}
S.~G. Mallat and Z.~Zhang, ``Matching pursuits with time-frequency
  dictionaries,'' \emph{Signal Processing, IEEE Transactions on}, vol.~41,
  no.~12, pp. 3397--3415, 1993.

\bibitem{pati1993orthogonal}
Y.~C. Pati, R.~Rezaiifar, and P.~Krishnaprasad, ``Orthogonal matching pursuit:
  Recursive function approximation with applications to wavelet
  decomposition,'' in \emph{Signals, Systems and Computers, 1993. 1993
  Conference Record of The Twenty-Seventh Asilomar Conference on}.\hskip 1em
  plus 0.5em minus 0.4em\relax IEEE, 1993, pp. 40--44.

\bibitem{chen1998atomic}
S.~S. Chen, D.~L. Donoho, and M.~A. Saunders, ``Atomic decomposition by basis
  pursuit,'' \emph{SIAM journal on scientific computing}, vol.~20, no.~1, pp.
  33--61, 1998.

\bibitem{aharon2006img}
M.~Aharon, M.~Elad, and A.~Bruckstein, ``K-svd: An algorithm for designing
  overcomplete dictionaries for sparse representation,'' \emph{Signal
  Processing, IEEE Transactions on}, vol.~54, no.~11, pp. 4311--4322, 2006.

\bibitem{lewicki2000learning}
M.~S. Lewicki and T.~J. Sejnowski, ``Learning overcomplete representations,''
  \emph{Neural computation}, vol.~12, no.~2, pp. 337--365, 2000.

\bibitem{mairal2012task}
J.~Mairal, F.~Bach, and J.~Ponce, ``Task-driven dictionary learning,''
  \emph{Pattern Analysis and Machine Intelligence, IEEE Transactions on},
  vol.~34, no.~4, pp. 791--804, 2012.

\bibitem{zhang2010discriminative}
Q.~Zhang and B.~Li, ``Discriminative k-svd for dictionary learning in face
  recognition,'' in \emph{Computer Vision and Pattern Recognition (CVPR), 2010
  IEEE Conference on}.\hskip 1em plus 0.5em minus 0.4em\relax IEEE, 2010, pp.
  2691--2698.

\bibitem{zheng2011graph}
M.~Zheng, J.~Bu, C.~Chen, C.~Wang, L.~Zhang, G.~Qiu, and D.~Cai, ``Graph
  regularized sparse coding for image representation,'' \emph{Image Processing,
  IEEE Transactions on}, vol.~20, no.~5, pp. 1327--1336, 2011.

\bibitem{gao2013laplacian}
S.~Gao, I.~W.-H. Tsang, and L.-T. Chia, ``Laplacian sparse coding, hypergraph
  laplacian sparse coding, and applications,'' \emph{IEEE Transactions on
  Pattern Analysis and Machine Intelligence}, pp. 92--104, 2013.

\bibitem{elad2006image}
M.~Elad and M.~Aharon, ``Image denoising via sparse and redundant
  representations over learned dictionaries,'' \emph{Image Processing, IEEE
  Transactions on}, vol.~15, no.~12, pp. 3736--3745, 2006.

\bibitem{yu2010improved}
K.~Yu and T.~Zhang, ``Improved local coordinate coding using local tangents,''
  in \emph{Proc. of the Int’l Conf. on Machine Learning (ICML)}, 2010.

\bibitem{zhang2011learning}
Z.~Zhang, L.~Ladicky, P.~Torr, and A.~Saffari, ``Learning anchor planes for
  classification,'' in \emph{Advances in Neural Information Processing
  Systems}, 2011, pp. 1611--1619.

\bibitem{huang2011salient}
Y.~Huang, K.~Huang, Y.~Yu, and T.~Tan, ``Salient coding for image
  classification,'' in \emph{Computer Vision and Pattern Recognition (CVPR),
  2011 IEEE Conference on}.\hskip 1em plus 0.5em minus 0.4em\relax IEEE, 2011,
  pp. 1753--1760.

\bibitem{sac11}
L.~Liu, L.~Wang, and X.~Liu, ``{In Defense of Soft-assignment Coding},'' 2011.

\bibitem{bo2009efficient}
L.~Bo and C.~Sminchisescu, ``Efficient match kernel between sets of features
  for visual recognition,'' \emph{Advances in neural information processing
  systems}, vol.~2, no.~3, 2009.

\bibitem{lin2011large}
Y.~Lin, F.~Lv, S.~Zhu, M.~Yang, T.~Cour, K.~Yu, L.~Cao, and T.~Huang,
  ``Large-scale image classification: fast feature extraction and svm
  training,'' in \emph{Computer Vision and Pattern Recognition (CVPR), 2011
  IEEE Conference on}.\hskip 1em plus 0.5em minus 0.4em\relax IEEE, 2011, pp.
  1689--1696.

\bibitem{xie2010large}
B.~Xie, M.~Song, and D.~Tao, ``Large-scale dictionary learning for local
  coordinate coding,'' in \emph{Proceedings of the British Machine Vision
  Conference (BMVC)}, 2010, pp. 36--1.

\bibitem{bishop2006pattern}
C.~M. Bishop \emph{et~al.}, \emph{Pattern recognition and machine
  learning}.\hskip 1em plus 0.5em minus 0.4em\relax springer New York, 2006,
  vol.~1.

\bibitem{zhu2005semi}
X.~Zhu, ``Semi-supervised learning literature survey,'' 2005.

\bibitem{belkin06}
M.~Belkin, P.~Niyogi, and V.~Sindhwani, ``Manifold regularization: A geometric
  framework for learning from labeled and unlabeled examples,'' \emph{The
  Journal of Machine Learning Research}, vol.~7, pp. 2399--2434, 2006.

\bibitem{jaakkola2002partially}
M.~S.~T. Jaakkola, ``Partially labeled classification with markov random
  walks,'' in \emph{Advances in Neural Information Processing Systems 14:
  Proceedings of the 2002 Conference}, vol.~2.\hskip 1em plus 0.5em minus
  0.4em\relax MIT Press, 2002, p. 945.

\bibitem{lafon05}
R.~Coifman, S.~Lafon, A.~Lee, M.~Maggioni, B.~Nadler, F.~Warner, and S.~Zucker,
  ``Geometric diffusions as a tool for harmonic analysis and structure
  definition of data: Diffusion maps,'' \emph{Proceedings of the National
  Academy of Sciences of the United States of America}, vol. 102, no.~21, p.
  7426, 2005.

\bibitem{me2013}
A.~Shaban, H.~R. Rabiee, M.~Farajtabar, and M.~Ghazvninejad, ``From local
  similarity to global coding; an application to image classification,'' in
  \emph{Computer Vision and Pattern Recognition (CVPR), IEEE Conference
  on}.\hskip 1em plus 0.5em minus 0.4em\relax IEEE, 2013.

\bibitem{lerman07}
G.~Lerman and B.~Shakhnovich, ``Defining functional distance using manifold
  embeddings of gene ontology annotations,'' \emph{Proceedings of the National
  Academy of Sciences}, vol. 104, no.~27, p. 11334, 2007.

\bibitem{singer09}
A.~Singer, R.~Erban, I.~Kevrekidis, and R.~Coifman, ``Detecting intrinsic slow
  variables in stochastic dynamical systems by anisotropic diffusion maps,''
  \emph{Proceedings of the National Academy of Sciences}, vol. 106, no.~38, pp.
  16\,090--16\,095, 2009.

\bibitem{l1graph13}
B.~Cheng, J.~Yang, S.~Yan, Y.~Fu, and T.~S. Huang, ``Learning with
  $\ell1$-graph for image analysis,'' \emph{Image Processing, IEEE Transactions
  on}, vol.~19, no.~4, pp. 858--866, 2010.

\bibitem{uci10}
\BIBentryALTinterwordspacing
A.~Frank and A.~Asuncion, ``{UCI} machine learning repository,'' 2010.
  [Online]. Available: \url{http://archive.ics.uci.edu/ml}
\BIBentrySTDinterwordspacing

\bibitem{libsvm11}
C.-C. Chang and C.-J. Lin, ``{LIBSVM}: A library for support vector machines,''
  \emph{ACM Transactions on Intelligent Systems and Technology}, vol.~2, pp.
  27:1--27:27, 2011, software available at
  \url{http://www.csie.ntu.edu.tw/~cjlin/libsvm}.

\bibitem{liblinear08}
R.-E. Fan, K.-W. Chang, C.-J. Hsieh, X.-R. Wang, and C.-J. Lin, ``{LIBLINEAR}:
  A library for large linear classification,'' \emph{Journal of Machine
  Learning Research}, vol.~9, pp. 1871--1874, 2008.

\bibitem{shi00}
J.~Shi and J.~Malik, ``Normalized cuts and image segmentation,'' \emph{IEEE
  Transactions on Pattern Analysis and Machine Intelligence}, vol.~22, no.~8,
  2000.

\bibitem{lovasz}
L.~Lov\'asz and M.~Plummer, \emph{Matching Theory}.\hskip 1em plus 0.5em minus
  0.4em\relax North-Holland, Amsterdam, 1986.

\end{thebibliography}
%




%

\begin{IEEEbiographynophoto}{Amirreza Shaban}
(S'12) received the B.Sc. degree from the School of Electrical and Computer Engineering at University of Tehran, Iran in 2010 and M.Sc. degree in computer engineering from Sharif University of Technology, Iran 2012. He is currently pursuing the Ph.D. degree at Ohio State University.

Since Fall 2013, he has been with the Department of Computer Science and Engineering at the Ohio State University. He is currently working with Prof. M. Belkin on his Ph.D. degree. His research interests include machine learning, image processing and computer vision.
\end{IEEEbiographynophoto}

\begin{IEEEbiography}[{\includegraphics[width=1.1in,height=1.35in,clip,keepaspectratio]{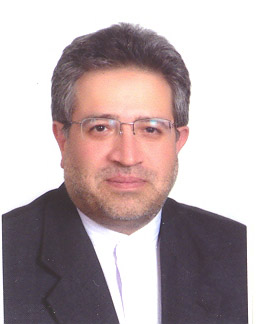}}]{Hamid R. Rabiee}
(SM'07) received his B.S. and M.S. degrees in Electrical Engineering from CSULB, USA, his EEE in Electrical and Computer Engineering from USC, USA and his Ph.D. in Electrical and Computer Engineering from Purdue University, West Lafayette, USA in 1996. From 1993 to 1996 he was a Member of Technical Staff at AT\&T Bell Laboratories. From 1996 to 1999 he worked as a Senior Software Engineer at Intel Corporation. He was also with PSU, OGI and OSU universities as an adjunct professor of Electrical and Computer Engineering from 1996 to 2000. Since September 2000, he has joined Sharif University of Technology, Tehran, Iran. He is the founder of Sharif Universitys Advanced Information and Communication Technology Research Center, Sharif University Advanced Technologies Incubator, Sharif Digital Media Laboratory and Mobile Value Added Services laboratories. He is currently an Professor of the Computer Engineering at Sharif University of Technology. He has been the initiator and director of national and international level projects in the context of UNDP International Open Source Network (IOSN) and Iran's National ICT Development Plan. He has received numerous awards and honors for his Industrial, scientific and academic contributions, and has acted as chairman in a number of national and international conferences, and holds three patents. He is also a Senior Member of IEEE.
\end{IEEEbiography}


\begin{IEEEbiographynophoto}{Mahyar Najibi}
received his B.Sc. degree from the School of Computer Engineering at Iran University of Science and Technology in 2011 and studied industrial engineering as his second B.Sc. major.
He received his M.Sc. degree in artificial intelligence from the School of Computer Engineering at Sharif University of Technology, Iran 2013. He is currently a member of the Sparse Signal Processing laboratory at Sharif University of Technology. His research interests are mainly focused on machine learning, computer vision and sparse signal processing.
\end{IEEEbiographynophoto}




\end{document}